\documentclass[preprint,11pt]{elsarticle}

\expandafter\let\csname equation*\endcsname\relax
\expandafter\let\csname endequation*\endcsname\relax
\usepackage{cancel}
\usepackage[dvipsnames,table]{xcolor}  
\usepackage{natbib}
\setcitestyle{square,numbers}
\usepackage{hyperref}
\newcommand\myshade{85}
\colorlet{mylinkcolor}{violet}
\colorlet{mycitecolor}{YellowOrange}
\colorlet{myurlcolor}{Aquamarine}
\hypersetup{
    colorlinks=true,
    linkcolor  = mylinkcolor!\myshade!black,
    citecolor  = mycitecolor!\myshade!black,
    urlcolor   = myurlcolor!\myshade!black,
}

\usepackage[shortlabels]{enumitem}
\usepackage{caption}

\usepackage{xargs}
\usepackage{subcaption}
\usepackage{multirow}
\usepackage{amsmath}
\usepackage{amssymb}
\usepackage{amsthm}
\usepackage{mathtools}
\usepackage{float}
\usepackage{jlcode}
\usepackage{nicefrac}
\usepackage{pifont}
\usepackage{booktabs}\let\cline\cmidrule
\usepackage{thm-restate}
\newtheorem{lemma}{Lemma}

\newtheorem{remark}{Remark}

\newcommand{\xmark}{\ding{55}}

\newcommand{\R}{\mathbb{R}}
\newcommand{\id}{\mathbb{I}}
\newcommand{\eqdef}{:=}
\newcommand{\params}{\mathbf{p}}
\newcommand{\trueParams}{\params^*}
\newcommand{\estim}{\hat{\params}}
\newcommand{\ustate}{\boldsymbol{u}}

\renewcommand{\t}{t}
\newcommand{\noise}{\boldsymbol{\varepsilon}}
\newcommand{\testFun}{\boldsymbol{\varphi}}
\newcommand{\rhs}{\boldsymbol{f}}
\newcommand{\dstate}{\mathbf{u}}
\newcommand{\dstatemat}{\mathbf{U}}
\newcommand{\dt}{\mathbf{t}}
\newcommand{\dTrueState}{\dstate^*}
\newcommand{\dTrueStateMat}{\dstatemat^*}
\newcommand{\dTestFun}{\boldsymbol{\Phi}}
\newcommand{\dRhs}{\mathbf{F}}
\newcommand{\dNoise}{\mathbf{\mathcal{E}}}
\newcommand{\wRhsLin}{\mathbf{G}}
\newcommand{\wRhs}{\mathbf{g}}
\newcommand{\wLhs}{\mathbf{b}}
\newcommand{\wRes}{\mathbf{r}}
\newcommand{\wCov}{\mathbf{S}}

\newcommand{\wResNoise}{\wRes^{\noise}}
\newcommand{\wResLin}{\wRes^{\mathrm{lin}}}
\newcommand{\wResQuad}{\wRes^\mathrm{int}}
\newcommand{\loglikelihood}{\ell}
\newcommand{\lognoise}{\boldsymbol{\nu}}
\newcommand{\logstate}{\boldsymbol{x}}
\newcommand{\dlognoise}{\mathbf{\mathcal{V}}}
\newcommand{\dlogstate}{\mathbf{x}}
\newcommand{\dlogstatemat}{\mathbf{X}}

\newcommand{\logRhs}{\boldsymbol{\tilde{f}}}
\newcommand{\wlogRhs}{\mathbf{\tilde{g}}}
\newcommand{\wlogLhs}{\mathbf{\tilde{b}}}
\newcommand{\wlogRes}{\mathbf{\tilde{r}}}
\newcommand{\wlogResLin}{\mathbf{\tilde{r}^{\mathrm{lin}}}}
\newcommand{\wlogCov}{\mathbf{\tilde{S}}}
\newcommand{\orcidlogo}{\includegraphics[width=3mm]{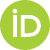}}
\newcommand{\orcid}[1]{\href{#1}{\orcidlogo}}

\journal{Applied Mathematics and Computation}

\begin{document}

\begin{frontmatter}
\title{Weak form Estimation of Nonlinear Dynamics (WENDy) for Nonlinear-in-Parameters ODEs}

\author[1]{Nicholas Rummel \orcid{https://orcid.org/0009-0005-2943-6953}\corref{cor1}} \ead{nicholas.rummel@colorado.edu}
\author[1]{Stephen Becker \orcid{https://orcid.org/0000-0002-1932-8159}}
\author[2]{Daniel Messenger \orcid{https://orcid.org/0000-0002-8275-7888}} 
\author[1]{Vanja Dukic \orcid{https://orcid.org/0000-0002-0348-0834}}
\author[1]{David Bortz \orcid{https://orcid.org/0000-0003-1163-7317}} \ead{david.bortz@colorado.edu}
\affiliation[1]{
    organization={Department of Applied Mathematics,
        University of Colorado},
    addressline={11 Engineering Drive},
    city={Boulder},
    postcode={80309},
    state={Colorado},
    country={USA}
}
\affiliation[2]{
    organization={Theoretical Division, 
        Los Alamos National Laboratory},
    city={Los Alamos}, 
    postcode={87545}, 
    state={New Mexico}, 
    country={USA}  
}
\cortext[cor1]{Corresponding author}



\begin{abstract}
    The Weak-form Estimation of Non-linear Dynamics (WENDy) framework is a recently developed approach for parameter estimation and inference of systems of ordinary differential equations (ODEs). Prior work demonstrated WENDy to be robust, computationally efficient, and accurate, but only works for ODEs which are linear-in-parameters. In this work, we derive a novel extension to accommodate systems of a more general class of ODEs that are nonlinear-in-parameters. Our new WENDy-MLE algorithm approximates a maximum likelihood estimator via local non-convex optimization methods. This is made possible by the availability of analytic expressions for the likelihood function and its first and second order derivatives. WENDy-MLE has better accuracy, a substantially larger domain of convergence, and is often faster than other weak form methods and the conventional output error least squares method. Moreover, we extend the framework to accommodate data corrupted by multiplicative log-normal noise. 

    The \href{https://github.com/nrummel/WENDy.jl}{WENDy.jl} algorithm is efficiently implemented in Julia. In order to demonstrate the practical benefits of our approach, we present extensive numerical results comparing our method, other weak form methods, and output error least squares on a suite of benchmark systems of ODEs in terms of accuracy, precision, bias, and coverage. 
\end{abstract}






\begin{highlights}
    \item Significantly increased domain of convergence 
    \item Derived the likelihood of weak-form equation error for multiplicative log-normal noise
    \item Successful estimation of nonlinear in parameter ODEs including Hill coefficients
\end{highlights}
\end{frontmatter}

\section{Introduction}

Parameter estimation and inference for differential equations (DEs) remains a challenging problem despite the many sophisticated algorithms that have been developed over the years \citep{McGoffMukherjeePillai2015StatSurv}. Precursors to modern approaches can be traced back to the dawn of computational science in the 1950's and 1960's \citep{Greenberg1951NACATN2340} and have been historically categorized into Output Error (OE) or Equation Error (EE) approaches \citep{Ljung1999,Ljung2017WileyEncyclopediaofElectricalandElectronicsEngineering}.
 
The more common OE-based methods first propose a candidate parameter value, numerically compute an approximate solution, compare with data, and then iterate the parameter proposal until the comparison metric drops below a specified threshold. OE methods have come to dominate the parameter estimation literature partially because they can leverage existing numerical simulation and nonlinear optimization approaches. However, the cost functions commonly resulting from nonlinear differential equation systems often exhibit multiple modes and ridges, and thus convergence to globally optimal and unique parameter values frequently requires long run times and careful diagnostic monitoring \citep{DukicLopesPolson2012JAmStatAssoc,KennedyDukicDwyer2014AmNat,RamsayHookerCampbellEtAl2007JRStatSocSerBStatMethodol, ZhangLuLiuEtAl2021NonlinearDyn}, in addition to the assessment of the  impact of hyperparameter choices from the DE solver and optimization algorithm  on the parameter estimates themselves \citep{NardiniBortz2019InverseProbl}. These  practical challenges have motivated alternative approaches with reduced reliance on forward-based solvers, including likelihood-free sequential Monte Carlo methods \citep{ToniWelchStrelkowaEtAl2009JRSocInterface}, local approximations \citep{ConradMarzoukPillaiEtAl2016JAmStatAssoc}, manifold-constrained Gaussian processes \citep{WongYangKou2024JStatSoft,YangWongKou2021ProcNatlAcadSciUSA}, and weak-form methods \citep{BortzMessengerDukic2023BullMathBiol,BrunelClairondAlche-Buc2014JAmStatAssoc,HallMa2014JRStatSocB}. 
 
EE-based estimation, on the other hand, involves substituting data directly into a model and minimizing the norm of the resulting residual. If the model equation is linear in the parameters (LiP), the resulting problem corresponds to a linear regression. This category also encompasses a variety of extensions that minimize transformed versions of the residual. For example, a Fourier Transform of the model equations results in a Fourier Error (FE) method. FE-methods are efficient, but that efficiency is (generally) limited to models that are LiP. Conversely, if the model is \emph{nonlinear} in the parameters (NiP), this results in a nonlinear regression problem, necessitating an iterative scheme for the optimization. The NiP EE-based scenario has not received nearly as much attention as the LiP case, forming a major motivation for our efforts here.

\subsection{Weak Form System Identification}

Another EE extension involves converting the model equation to its weak form, i.e., convolving with a compactly supported test function, $\varphi$, and integrating by parts. This approach was originally proposed in 1954 \cite{Shinbrot1954NACATN3288}, rediscovered in 1965 \cite{LoebCahen1965IEEETransAutomControl,LoebCahen1965IEEETransAutomControl}, and came to be known as the \emph{Modulating Function Method} (MFM) in the control engineering literature. Identical ideas have also been (independently) explored in statistics \citep{BrunelClairondAlche-Buc2014JAmStatAssoc,HallMa2014JRStatSocB} and nonlinear dynamics \citep{BortzMessengerDukic2023BullMathBiol,LiuChangChen2016NonlinearDyn}. Over the years researchers have proposed several forms for $\varphi$ including those based upon piecewise \citep{LoebCahen1965IEEETransAutomControl} and Hermite \citep{Takaya1968IEEETransAutomControl} polynomials, Fourier \citep{PearsonLee1985Control-TheoryAdvTechnol} and Hartley \citep{PatraUnbehauen1995IntJControl} Transforms, and Volterra linear integral operators \citep{PinAssaloneLoveraEtAl2015IEEETransAutomatContr,PinChenParisini2017Automatica}.  However, while it was known that the choice of $\varphi$ impacts the estimation and inference accuracy, only recently have there been precise derivations of the impact of test function hyperparameters on the solution \cite{BortzMessengerDukic2023BullMathBiol,GurevichReinboldGrigoriev2019Chaos,MessengerBortz2021JComputPhys,MessengerBortz2021MultiscaleModelSimul,MessengerBortz2024IMAJNumerAnal,WangLiuGibaru2023NonlinearDyna}.

Starting in 2010, research in two independent areas led to a resurgence in interest in weak form system identification. Janiczek noted that a generalized version of integration-by-parts could extend the traditional MFM to work with fractional differential equations \citep{Janiczek2010BullPolAcadSciTechSci}. This discovery led to a renewed (and sustaining) interest by several control engineering researchers, c.f.~\citep{AldoghaitherLiuLaleg-Kirati2015SIAMJSciComput,JouffroyReger20152015IEEEConfControlApplCCA,LiuLaleg-Kirati2015SignalProcessing}.

The second independent line of weak form research arose out of the field of sparse regression-based equation learning. While not a weak form method, the seminal work in this area is the Sparse Identification of Nonlinear Dynamics (SINDy) methodology and was originally proposed in 2016 for ODEs \citep{BruntonProctorKutz2016ProcNatlAcadSci} and in 2017 for PDEs \citep{RudyBruntonProctorEtAl2017SciAdv,Schaeffer2017ProcRSocMathPhysEngSci}. It is a data-driven modeling approach that learns both the form of the differential equation and the parameters directly from data. While incredibly successful, the original SINDy method was a strong form method and thus not robust to noise. Subsequently, several researchers independently noticed the value in casting equations in the weak form \citep{GurevichReinboldGrigoriev2019Chaos,MessengerBortz2021JComputPhys,MessengerBortz2021MultiscaleModelSimul,PantazisTsamardinos2019Bioinformatics,WangHuanGarikipati2019ComputMethodsApplMechEng}. In particular, we note that the Weak form Sparse Identification of Nonlinear Dynamics (WSINDy) method proposed in  \citep{MessengerBortz2021JComputPhys,MessengerBortz2021MultiscaleModelSimul} clearly demonstrates the noise robustness and computational efficiency of a weak form approach to equation learning. We direct the interested reader to recent overviews \cite{BortzMessengerTran2024NumericalAnalysisMeetsMachineLearning,MessengerTranDukicEtAl2024SIAMNews} and further WSINDy extensions \cite{MessengerBortz2022PhysicaD,MessengerBurbyBortz2024SciRep,MessengerDallAneseBortz2022ProcThirdMathSciMachLearnConf,MessengerWheelerLiuEtAl2022JRSocInterface,RussoMessengerBortzEtAl2024IFAC-PapersOnLine,TranHeMessengerEtAl2024ComputMethodsApplMechEng}.

Lastly, we note that all the above EE-based approaches rarely address the fact that the data appears in terms on both sides of the EE regression equation, i.e., in the matrix and the vector.  This class of problems is known as an Errors-in-Variables problem \citep{Durbin1954RevIntStatInst} and can be addressed using methods from the generalized least squares literature. In our previous work \citep{BortzMessengerDukic2023BullMathBiol}, we proposed using an iteratively reweighted least squares approach (IRLS) which repeatedly solves the EE regression, updating the covariance at each step. In what follows, we denote the original WENDy algorithm as WENDy-IRLS.

\subsection{Contribution}
The current work builds on the analytic framework in \citep{BortzMessengerDukic2023BullMathBiol}, providing a more precise understanding of the approximate distribution of the weak residual and extending the framework to a broader class of ODEs that have a nonlinear dependence on the parameters to be estimated. Furthermore, the framework of our derivation applies beyond additive Gaussian noise, and now can explicitly address multiplicative log-normal noise. The distribution is leveraged in our novel algorithm, called WENDy-MLE, which significantly improves on the previous WENDy-IRLS algorithm presented in \citep{BortzMessengerDukic2023BullMathBiol}. The algorithm is based on likelihood function optimization and designed to find the maximum likelihood estimate of the parameters. This is done via an efficient implementation in Julia \citep{BezansonEdelmanKarpinskiEtAl2017SIAMRev} with analytic computation of first and second order derivatives of the likelihood, thus allowing users to easily apply it to their ODE. Through numerical results, we demonstrate that the algorithm converges to accurate estimates of the parameters more often than a conventional forward solver-based least squares method.  

The paper is structured as follows: Section \ref{sec:mamathematical-frameworkth} describes the mathematical derivation of the WENDy framework. Section \ref{sec:results} presents a brief overview of the implementation details and numerical results comparing the WENDy-MLE algorithm to other existing methods. We conclude this work in Section \ref{sec:discussion} with a discussion of strengths and weaknesses in our approach including how our approach compares with other state-of-the-art methods. Concluding remarks are made in Section \ref{sec:conclusion}, and acknowledgments are made in Section \ref{sec:ack}.
\section{WENDy Mathematical Framework} \label{sec:mamathematical-frameworkth}
This section lays out the mathematical foundation and needed assumptions for the new WENDy-MLE algorithm. It first describes the class of differential equations and noise for which the algorithm is applicable. We then present the approximate distribution of the weak residual and the corresponding likelihood function. 
Finally, we extend the framework to the case of multiplicative log-normal noise. We provide additional details, full proofs, analytic expressions for derivatives of the weak likelihood, and analytic simplifications that lead to computational benefits in \ref{appendix:TestFuns}, \ref{appendix:proofs}, \ref{appendix:DerInfo}, and \ref{appendix:linvsnonlin} respectively.

To set notation, we assume that the dynamical system is governed by a fully-observed $D$-dimensional system of ordinary differential equations (ODE) of the form 
\begin{equation}
	\label{eq:ode-strong}
	\dot{\ustate}(\t) = \rhs(\params,\ustate(\t),\t)
\end{equation}
where $\ustate(\t)$
is the function state variable at time $\t \in [0,T]$; we typically drop the $\t$ dependence and write $\dot{\ustate} = \rhs(\params, \ustate,\t)$ for short. 
The driving function $\rhs$ is (possibly nonlinearly) parameterized by a finite set of unknown parameters $\params\in\R^J$ that we wish to recover, and $\rhs$ may also be nonlinear in $\ustate$ and $\t$. It is required that $\rhs$ be twice continuously differentiable with respect to $\params$ and  $\ustate$. Note that we use bold lowercase variables to indicate vectors and vector-valued functions, and bold upper case to indicate matrices and matrix-valued functions. We use $\|\cdot\|$ to denote the 2-norm for vectors and the Frobenius norm for matrices unless otherwise specified.

\subsection{The Weak Form}

To implement the WENDy-MLE approach, we first convert Equation \eqref{eq:ode-strong} from strong to weak form. This is done by taking the $L^2\bigl((0,T)\bigr)$ inner product, $\langle \cdot, \cdot\rangle$, with an analytic compactly supported test function, $\testFun\in \mathcal{C}^\infty_c\bigl((0,T), \R^D\bigr)$. In this work, multivariate test functions are chosen to be of the form $\testFun(\t) = \mathbf{1}_D \varphi(t)$ where $\varphi \in \mathcal{C}^\infty_c\bigl((0,T), \R\bigr)$\footnote{One could choose more general test functions $\testFun(t) = [\varphi_1(t), \cdots, \varphi_D(t)]^\top$ and the derivations here can be easily extended to this case. This is also a slight abuse of notation. The inner product $\langle \cdot, \cdot \rangle$ when applied to $D$ dimensional functions produces a $D$ dimensional vector rather than a scalar.}. The weak form now becomes
\begin{equation}
	\label{eq:ode-weak}
	\langle \testFun, \dot{\ustate}\rangle = \langle \testFun,\rhs(\params, \ustate, \t)\rangle.
\end{equation}
Because the test functions have compact support, integration by parts allows movement of the derivatives onto the test functions: 
\begin{equation}
	\label{eq:ode-weak-trick}
	-\langle \dot{\testFun}, \ustate\rangle = \langle \testFun, \rhs(\params, \ustate, \t)\rangle.
\end{equation}

\subsection{Description of Data with Gaussian Noise} \label{sec:dataDesc}
For the data generating mechanism, we assume that the  observations arise when the solution to a system of the form \eqref{eq:ode-strong} gets corrupted by noise. For simplicity, the data is assumed to be observed on a uniform grid with stepsize $\Delta \t$. For the Gaussian case, we further assume that the noise is additive, consisting of independent and identically distributed (i.i.d.) Gaussian random variables. Specifically, for sample timepoints $\{t_m\}_{m=0}^M$, and true states $\dTrueState_m:=\ustate(t_m)$, we assume that the observed data arise as  a sum of the true state and noise: 
$$\dstate_m = \dTrueState_m+\noise_m\quad \forall m \in \{0, ..., M\}$$ 
where $\noise_m \stackrel{iid}{\sim}\mathcal{N}({\mathbf 0},{\mathbf{\Sigma}})$, and ${\mathbf{\Sigma}}$ is a $D \times D$ diagonal matrix. Thus, the observed data is given as $\dstatemat = \dTrueStateMat + \dNoise$, where all matrices are in $\R^{(M+1) \times D}$. In Section \ref{sec:log} we will adapt this framework to accommodate the multiplicative log-normal noise instead of Gaussian. 

\subsection{Discretization}
Formally, to satisfy the weak form, a solution must solve the Equation \eqref{eq:ode-weak} (and equivalently Equation \eqref{eq:ode-weak-trick}) for all possible test functions. However, estimation of the parameters in Eq.~\eqref{eq:ode-weak} is the primary goal, so a finite set of test functions, $\{\testFun_k\}_{k=1}^K$, are chosen to best preserve the information of the data while suppressing the effects due to noise. The strategy for choosing an optimal set of test functions for a specific problem is an area of ongoing research. Here, the test function selection mirrors that described in \citep{BortzMessengerDukic2023BullMathBiol} and full details are available in \ref{appendix:TestFuns}. 

For the set of test functions $\{\varphi_k\}_{k=1}^K$, we build the matrices 
\[
    \dTestFun = \begin{bmatrix}
        \varphi_1(\dt)^\top \\ \vdots \\ \varphi_K(\dt)^\top
    \end{bmatrix}\in \R^{K\times (M+1)},\; \dot{\dTestFun} = \begin{bmatrix}
		\dot{\varphi}_1(\dt)^\top \\ \vdots \\ \dot{\varphi}_K(\dt)^\top
    \end{bmatrix}\in \R^{K\times (M+1)} .
\]
The matrices of the data and the RHS are as follows:
\[
	\dt \eqdef \begin{bmatrix}
		\t_0 \\ \vdots \\ \t_M
	\end{bmatrix}\in \R^{(M+1)\times 1}, \;
	\dstatemat \eqdef \begin{bmatrix}
		\dstate_0 ^\top\\
		\vdots \\
		\dstate_M^\top
	\end{bmatrix}\in\R^{(M+1) \times D},
\]
\[
	\dRhs(\params;\dstatemat,\dt) \eqdef \begin{bmatrix}
		\rhs(\params, \dstate_1,\t_1)^\top \\
		\vdots \\
		\rhs(\params, \dstate_M,\t_M)^\top 
	\end{bmatrix}\in\R^{(M+1) \times D}.
\]
The products $\dot{\dTestFun}\dstatemat$ and $\dTestFun\dRhs$ are equivalent to using the trapezoid rule to numerically approximate the integral in the inner product\footnote{If the grid was non-uniform or if a different quadrature rule was desired, then the test function matrices would be post-multiplied by a quadrature matrix $\mathcal{Q}$.  In the case of the trapezoid rule, that matrix is $\Delta t I$ except for the $(1,1)$ and $(M+1,M+1)$ entries which are $\nicefrac{\Delta t}{2}$. However, the compact support of $\varphi$ means that the $\nicefrac{\Delta t}{2}$ values will be multiplied by a zero and thus we can disregard the whole matrix (see the discussion between Equations (5) and (6) on page 7 of \citep{BortzMessengerDukic2023BullMathBiol}).}. Therefore, the discretization of \eqref{eq:ode-weak-trick} is
\begin{equation}
	\label{eq:weakSys}
	-\dot{\dTestFun} \dstatemat \approx \dTestFun \dRhs(\params;\dstatemat,\dt)
\end{equation}
and would be satisfied perfectly if there was no noise and no numerical error in the quadrature. This motivates inspecting the weak residual, which we define as: 
\begin{equation}
	\label{eq:weakRes}
	\wRes(\params) \eqdef \wRhs(\params;\dstatemat,\dt) - \wLhs(\dstatemat)
\end{equation}
where 
\[
\wRhs(\params;\dstatemat,\dt) \eqdef \operatorname{vec}[ \dTestFun \dRhs(\params;\dstatemat, \dt)], \;
\wLhs(\dstatemat) \eqdef - \operatorname{vec}[\dot{\dTestFun} \dstatemat]
\]
and ``$\operatorname{vec}$'' is the columnwise vectorization of a matrix. Since the noise $\noise_m \stackrel{iid}{\sim} N({\mathbf{0}},{\mathbf{\Sigma}})$, we know that  $\operatorname{vec}[\dNoise] \sim N({\mathbf{0}},  \mathbf{\Sigma} \otimes \id_{M+1})$ where $\otimes$ is the Kronecker product and $\id_{M+1}\in \mathbb{R}^{(M+1)\times (M+1)}$ is the identity matrix. 

\subsection{An Approximate Distribution of the Weak Residual}

The idea of using the weak form equation error residual has been used since the 1950's for parameter estimation \citep{Shinbrot1954NACATN3288}. However, equation error-based estimation (both weak and strong form) exhibit a known bias \citep{Chesher1991Biometrika}. By treating the weak residual as a random variable and approximating its distribution, the WENDy framework improves the quality of the parameter estimates.

In this section, the key theoretical result of our paper is stated in Proposition \ref{prop:wres-dist}, followed by several remarks on the consequences of this result. This derivation is a natural extension of the one in \citep{BortzMessengerDukic2023BullMathBiol}. The key difference is that now it is not assumed that $\dRhs$ is linear in $\params$, and $\dRhs$ can explicitly depend on $\t$. 

This result is derived by decoupling the numerical integration error from contribution of the noise. Then, by linearizing about the data we can define an approximation of the weak residual, denoted $\wResLin$. The distribution of $\wResLin$ is exactly multivariate Gaussian with covariance $\wCov$. The precise definitions of $\wResLin$ and $\wCov$ as well as the full proofs of Proposition \ref{prop:wres-dist} and its supporting Lemmas are in \ref{appendix:proofs}. 

\begin{restatable}{proposition}{primeProb} \label{prop:wres-dist}
	Let uncorrupted data $\dTrueStateMat$ and true parameters $\trueParams$ satisfy Equation \eqref{eq:ode-weak} on the time domain $[0,T]$. For   corrupted data $\dstatemat = \dTrueStateMat + \dNoise$ with  noise $\operatorname{vec}[\dNoise] \sim \mathcal{N}(\mathbf{0},  \mathbf{\Sigma} \otimes \id_{M+1})$, and $\rhs$ that is continuous in $\t$ and twice continuously differentiable in $\ustate$, the following holds: 
	\[ \wCov(\trueParams)^{-\tfrac{1}{2}}\wResLin(\trueParams) \sim \mathcal{N}({\mathbf{0}}, \id_{KD}) \;
	\textit{and} \;
	\lim_{M\rightarrow\infty} \mathbb{E}\Bigl[\bigl\|\wRes(\trueParams) - \wResLin(\trueParams)\bigr\|\Bigr] = \mathcal{O}\Bigl(\mathbb{E}\bigl[\|\dNoise\|^2\bigr]\Bigr).\]
\end{restatable}

\begin{remark}[Spectral Convergence of Integration Error]
	The grid is assumed to be uniform, so the quadrature rule chosen is the trapezoidal rule. This quadrature rule has spectral convergence because the periodic extension of a compact functions is itself \citep{Atkinson1989}. In particular, the order of the numerical error is $\mathcal{O}(\Delta t)^{(p+1)}$ where $p$ is the order of the smoothness of the test function at the endpoints of the integral domain. Thus in this context, error from numerical integration is negligible due the smoothness of $\varphi \ast u$. 
\end{remark}

\begin{remark}[Higher Order Terms]
    Unfortunately, higher order terms may not be negligible in noise regimes of interest, making the likelihood from the distribution in Proposition \ref{prop:wres-dist}  unrepresentative of the true likelihood of the weak residual. In practice, the Gaussian approximation still leads to reasonable estimates of the parameters, especially when $\rhs$ has less nonlinearity with respect to $\ustate$ and at lower noise levels. 
\end{remark}
\begin{remark}[Intuition]
    By linearizing the weak residual $\wRes$ about the state variable $\ustate$, the Gaussianity of the noise is preserved. Linear combinations of Gaussians remain Gaussian.
\end{remark}

\subsection{Weak Likelihood Function}
In \citep{BortzMessengerDukic2023BullMathBiol}, the (unweighted) weak form least squares (WLS) was improved by utilizing covariance information in the WENDy-IRLS algorithm. Experimentally, it was found that WENDy-IRLS convergence degrades when differential systems were highly nonlinear in the state variable $\ustate$ and when noise levels became too high. To address these concerns consider optimizing the negative log-likelihood of the weak residual to approximate the MLE. Because the weak residual is approximately distributed multivariate normal, the analytic form of the negative log-likelihood is known. By plugging in the values for its mean and covariance the weak likelihood is obtained to be
\begin{equation}\label{eq:wnll}
	\loglikelihood(\params) = \operatorname{logdet}\bigl(\wCov(\params)\bigr) + \wRes(\params)^\top\wCov(\params)^{-1}\wRes(\params) .
\end{equation}
Furthermore, analytic gradient and Hessian information of \eqref{eq:wnll} can be efficiently computed. Thus, we can make use of modern optimization algorithms to find the MLE. A derivation of derivative information is available in \ref{appendix:DerInfo}.

\begin{remark}[Implicit Regularization] 
Notice that $\loglikelihood$ is the sum of two terms. The first, $\operatorname{logdet}\bigl(\wCov(\params)\bigr)$, only depends on the eigenvalues of the covariance, and penalizes parameter values that lead to large uncertainty. The second term is the Mahalanobis Distance which is the Euclidean norm weighted by the inverse of the covariance matrix, and penalizes parameter values that lead to large residuals. The contribution of the second term acts as a regularizer, and balancing these last two terms minimizes uncertainty and equation error. 
\end{remark} 

\subsection{Extending to Log-Normal Noise} \label{sec:log}
In many systems, it is more appropriate to have multiplicative log-normal noise rather than additive Gaussian noise. This is particularly true when the state variable cannot be negative. This formally means the data is corrupted in the following way
\[\dstate_m = \dTrueState_m \circ \lognoise_m \quad \forall m \in \{0, \cdots, M\} \]
where \(\log(\lognoise_m) \stackrel{\text{iid}}{\sim} N({\mathbf{0}}, {\mathbf{\Sigma}})\) and $\circ$ denotes the Hadamard Product meaning multiplication occurs element-wise. Our convention is that logarithms and exponentials are applied element-wise.

The result from Proposition \ref{prop:wres-dist} can be extended to the log-normal case by applying a transformation to the state variable. For the transformed system of equations, we can define expressions for the weak residual $\wlogRes$, its linearization $\wlogResLin$, and its covariance $\wlogCov$. The precise definitions of these expressions as well as the full proof of Corollary \ref{cor:log-dist} are in \ref{appendix:proofs}. 

\begin{restatable}{corollary}{logCor} \label{cor:log-dist} 
	Let true data $\dTrueStateMat$ and true parameters $\trueParams$ satisfy Equation \eqref{eq:ode-weak} on the time domain $[0,T]$, with noise $\log(\dlognoise) \sim \mathcal{N}(\mathbf{0},  \mathbf{\Sigma} \otimes \id_{M+1})$, and corrupted data $\dstatemat = \dTrueStateMat \circ \dlognoise$. Assuming  $\logRhs$ is continuous in $\t$ and twice continuously differentiable in $\ustate$, the following holds for transformed weak residual:
	\[\wlogCov(\trueParams)^{-\tfrac{1}{2}} \wlogResLin(\trueParams) \sim N\bigl({\mathbf{0}}, \id_{KD}\bigr) \; \textit{and} \; \mathbb{E}\Bigl[\bigl\|\wlogRes(\trueParams) - \wlogResLin(\trueParams)\bigr\|\Bigr] = \mathcal{O}\Bigl(\mathbb{E}\bigl[\|\log(\dlognoise)\|^2\bigr]\Bigr)\]
\end{restatable}

\subsection{Approximate Distribution of WENDy-MLE Estimator}
Defining the maximum likelihood estimate for the approximate likelihood function defined in Equation \eqref{eq:wnll} to be 
\begin{equation} \label{eq:wendyOptim}
	\estim_\mathrm{mle} \eqdef \operatorname{argmin} \ell(\params)
\end{equation} 

Assuming $\estim_\mathrm{mle}$ can be found, then one can obtain an approximate distribution of the estimator
\begin{equation} \label{eq:param-dist}
    \begin{gathered}
	\mathbf{C}(\estim_\mathrm{mle})^{-\tfrac{1}{2}}\estim_\mathrm{mle} \stackrel{approx}{\sim} \mathcal{N}\bigl(\trueParams, \id_{J} \bigr) \\
	\text{ where }
	\mathbf{C}(\estim_\mathrm{mle}) = \nabla_{\params} \wRhs(\estim_\mathrm{mle})^{+}\wCov(\estim_\mathrm{mle})\bigl(\nabla_{\params} \wRhs(\estim_\mathrm{mle})^{+}\bigr)^\top.
	\end{gathered}
\end{equation}
The operator \((\cdot)^{+}\) is the Moore–Penrose inverse and $\nabla_{\params} \wRhs$ is the Jacobian of the $\wRhs$ with respect to $\params$. 

This is done via a linearization of the weak residual with respect to $\params$. This procedure is possible when the map $\wRhs(\cdot)$ is injective. We have found this to be true empirically as $\nabla_{\params} \wRhs(\estim_\mathrm{mle})$ is full column rank. 
This provides powerful information about estimated parameters; the algorithm is able to provide both an estimate and the uncertainty information about the estimate. 

\section{Implementation and Numerical Results}\label{sec:results}

This section contains a description of the maximum likelihood WENDy-MLE algorithm as well as the numerical results. In Section \ref{sec:wendyAlgo} we will discuss the algorithmic implementation. Then we introduce the details of comparison methods in Section \ref{sec:otherApproaches}. Next, the details of the metrics are shown in Section \ref{sec:metrics}, and artificial noise is discussed in Section \ref{sec:artificialNoise}. Our results begin in Section \ref{sec:motiv} where a motivating example demonstrates why alternatives to output error based methods are necessary. Finally in Section \ref{sec:mega}, results for a suite of test problems demonstrate key capabilities of the WENDy-MLE algorithm. 

\subsection{WENDy-MLE Algorithm} \label{sec:wendyAlgo}
Estimating the parameters can be framed the optimization problem specified in Equation~\eqref{eq:wendyOptim}. In general, this is a non-convex optimization problem, so it is appropriate to use trust-region second order methods that have been developed for this purpose. WENDy-MLE takes the time grid $\dt$, data $\dstatemat$, a function for the RHS $\rhs$, an initial guess for the parameters $\params_0$. Optionally, our algorithm accepts box constraints for the parameters $\params_\ell$ and $\params_u$ and solves the constrained optimization problem 
\begin{equation}\label{eq:wendyOptimConstrained}
	\estim_\mathrm{mle} = \underset{\substack{\params \in \mathbb{R}^J\\ \params_\ell \leq \params \leq \params_u }}{\operatorname{argmin}} \loglikelihood(\params).
\end{equation}

\href{https://docs.sciml.ai/Symbolics/stable/}{Symbolics.jl} \citep{GowdaMaCheliEtAl2021ACMCommunComputAlgebra} analytically computes derivatives of $\rhs$, and then Julia functions are formed from those symbolic expression. Then the weak likelihood and its derivatives can be evaluated using Equations \eqref{eq:wnll}, \eqref{eq:wnll-grad}, and \eqref{eq:wnll-hess}. This makes use of second order optimization methods possible\footnote{While the likelihood is a scalar valued function, its computation relies on the derivatives of vector and matrix valued functions. Building and using efficient data structures to compute these derivative can be done with ``vectorization'' resulting in large matrices with block structure from Kronecker products. In our implementation we instead use multidimensional arrays and define the operations in Einstein summation notation. These computations are then evaluated efficiently with \href{https://github.com/mcabbott/Tullio.jl?tab=readme-ov-file}{Tullio.jl} \citep{AbbottAluthgeN3N5EtAl2023}.}. Trust region solvers are provided by \href{https://github.com/JuliaSmoothOptimizers/JSOSolvers.jl}{JSOSolvers.jl} and \href{https://julianlsolvers.github.io/Optim.jl/stable/}{Optim.jl} for the constrained and unconstrained cases respectively \citep{KMogensenNRiseth2018JOSS, MigotOrbanSoaresSiqueira2024}. We note that our code also supports using the Adaptive Regularization Cubics variant (ARCqK) in the unconstrained case provided by \href{https://jso.dev/AdaptiveRegularization.jl/stable/} {AdaptiveRegularization.jl} \citep{DussaultMigotOrban2024MathProgram}. The trust region solvers and ARCqK usually produce similar results, but in our limited testing we found the trust region solvers work better in general, so only results for the trust region solver are shown. Our code \href{https://github.com/nrummel/WENDy.jl}{WENDy.jl} is readily available through the Julia package manager. Code to run the experiments and make the plots in this paper is available on \href{https://github.com/nrummel/rummelWENDy2025}{GitHub} as well.

\subsection{Fair Comparison With Other Approaches} \label{sec:otherApproaches}
We compare WENDy-MLE to a standard output error approach, as well as to the other weak form methods WLS and WENDy-IRLS. More details on the implementation and extension of WLS and WENDy-IRLS to the NiP case are found in \ref{appendix:otherWeak}.

When designing experiments, we took care in making fair comparisons by choosing hyper-parameterizations for each algorithm. All methods are given absolute and relative error tolerances of $10^{-8}$. This corresponds to the output error method having the same tolerances for its forward simulation of the system and the stopping criterion for optimization of the parameters. All solvers are said to have converged if the norm of the gradient, the norm of successive iterates, or the absolute value of the change in the loss is less than the tolerance in a relative or absolute sense. Furthermore, we limit all solvers maximum number of iterations to be 200 except for WENDy-IRLS where we set the maximum number of iterations to be 1000 because the computational cost of each iteration is much lower than for WENDy-MLE. We also set a run time limit of 200 seconds for all solvers. For the hyper-parameterizations stated, all solvers can complete their optimization in the time and iteration limits for most of the experiments. In particular, we have observed the output error solver's run time can explode when we require stricter tolerances in terms of the parameters that are being optimized. 

There are several existing state of the art methods which have been developed for estimating parameters of ODE's. Three of the most popular are 1) output error least squares, 2) collocation-based, and 3) Monte Carlo-based. We chose to single out OE-LS as it is the most ubiquitous tool used in the literature. And, we note that for the case of isotropic additive Gaussian noise the OE-LS approach is equivalent to maximizing the likelihood of forward simulated state variable\footnote{In the case of multiplicative log-normal noise, a transformation of the state variable similar to that used in Section \ref{sec:log} could be used to form a output likelihood maximization algorithm, but this was considered to be beyond the scope of this work.}. Collocation methods similar to that presented in \citep{RamsayHookerCampbellEtAl2007JRStatSocSerBStatMethodol} apply equation error regression over smoothed data. They are similar to the weak form methods presented here but require specific tuning of hyper-parameters while our weak form methods adaptively choose the test functions from the data alone. Monte Carlo-based methods are also widely used, but fall into a different (and highly computationally intensive) class of algorithms. They require tens of thousands (or more) forward simulations to generate samples, which leads to run times orders of magnitude longer than those presented here.

\subsubsection{Output Error Solver} \label{sec:fslsq}
For comparison, we use a state-of-the-art implementation of the standard output error approach of regressing the approximated state against the data. This is done by iteratively calling a direct numerical forward solver and optimizing over both the initial condition and the unknown parameters. We include the initial conditions as an optimization parameter since otherwise noisy initial data can drastically effect output error depending on the sensitivity of the system. The full scheme  can be framed as a nonlinear least squares problem:
\begin{equation}
	\label{eq:fslsq}
	\estim_\mathrm{oels} = \underset{\params \in \mathbb{R}^J }{\operatorname{argmin}} \Biggl[ \min_{\dstate_0 \in \mathbb{R}^D} \tfrac{1}{2} \Bigl\| \hat{\dstatemat}(\params, \dstate_0) - \dstatemat\Bigr\|^2_F \Biggr].
\end{equation}
We refer to this method Output Error Least Squares (OE-LS). Forward simulation is accomplished with the Rodas4P algorithm provided by \href{https://docs.sciml.ai/DiffEqDocs/stable/}{DifferentialEquations.jl} \citep{RackauckasNie2017JORS}. This solver is well suited to solve stiff system, so it is appropriate for all the example problems discussed here. Nonlinear least squares problems are solved by the Levenberg-Marquardt algorithm through \href{https://docs.sciml.ai/NonlinearSolve/stable/}{NonlinearSolve.jl} \citep{PalHoltorfLarssonEtAl2024arXiv240316341}. Derivative information is computed via automatic differentiation provided by \href{https://github.com/JuliaDiff/ForwardDiff.jl}{ForwardDiff.jl} \citep{RevelsLubinPapamarkou2016arXiv160707892}.

\subsubsection{Hybrid Algorithm} \label{sec:hybrid}
Combining equation error and output error methods can be advantageous. We found that using the WENDy-MLE algorithm to provide an initialization for OE-LS can lead to best results. The WENDy-MLE experimentally has been shown to have a larger domain of convergence, but fitting the trajectory of the dynamical system may still be the primary objective. If the weak likelihood evaluated at the parameters returned by the OE-LS method is 5\% worse than that of the weak likelihood of parameters returned by WENDy-MLE, then the method falls back to the estimate provided by WENDy-MLE.

\subsection{Metrics} \label{sec:metrics}
Depending on the motivation behind the system identification, one may either care about identifying the parameters themselves, or only care about predictive performance. Hence we look at two complementary accuracy metrics: relative coefficient error and relative forward simulation error. Furthermore, we provide statistical metrics for our estimator, including estimated relative bias, estimated relative variance, estimated relative mean square error, and coverage. All metrics are defined in Table \ref{table:metrics}.

\begin{table}[!ht]
	\resizebox{\textwidth}{!}{
    \begin{tabular}{lr}
            \toprule 
		  \textbf{Metric} &  \multicolumn{1}{c}{\textbf{Formula}}\\
		\midrule 
		\\[-5pt]
		Failure Rate & \begin{minipage}{7cm}
			\begin{equation*} \label{eq:failRate}\begin{gathered}
					\frac{|\mathcal{I}^c|}{N} \\
					\mathcal{I} = \Biggl\{n \mid \dfrac{\|\estim^{(n)} - \trueParams\|_2}{\|\trueParams\|_2} < 25, \\ \hspace{1cm}\dfrac{ \left\| \hat{\dstatemat}(\estim^{(n)}, \dstate_0^*) - \dTrueStateMat\right\|_F}{ \left\| \dTrueStateMat\right\|_F}  < 25 \Biggr\} \\[5pt]
			\end{gathered}
			\end{equation*}
		\end{minipage} \\
		\hline 
		\\[-5pt]
		Relative Coefficient Error & \begin{minipage}{7cm}
			\begin{equation*} \label{eq:cl2}
				\frac{1}{|\mathcal{I}|} \sum_{n \in \mathcal{I}} \dfrac{\|\estim^{(n)} - \trueParams\|_2}{\|\trueParams\|_2}\\[5pt]
			\end{equation*}
		\end{minipage} \\
		\hline 
		\\[-5pt]
		Relative Forward Simulation Error & \begin{minipage}{7cm}
			\begin{equation*} \label{eq:fsl2}
				\frac{1}{|\mathcal{I}|} \sum_{n \in \mathcal{I}} \dfrac{ \left\| \hat{\dstatemat}(\estim^{(n)}, \dstate_0^*) - \dTrueStateMat\right\|_F}{ \left\| \dTrueStateMat\right\|_F} \\[5pt]
			\end{equation*}
		\end{minipage} \\
        \hline  
		\\[-5pt]
		Estimated Relative Squared Bias & \begin{minipage}{7cm}
			\begin{equation*} \label{eq:estimBias}
				\underset{n\in \mathcal{I}}{\operatorname{med}}\left[\frac{\bigl(\hat{p}_j^{(n)} - p^*_j\bigr)^2}{(p^*_j)^2}  \right] \\[5pt]
			\end{equation*}
		\end{minipage} \\
		\hline 
		\\[-5pt]
		Estimated Relative Variance & \begin{minipage}{7cm}
			\begin{equation*} \label{eq:estimVar}
                \begin{gathered}
    				\underset{n\in \mathcal{I}}{\operatorname{med}}\left[\frac{\bigl(\hat{p}_j^{(n)} - \bar{p}_j\bigr)^2}{(p^*_j)^2}\right]\\
                    \bar{p}_j = \frac{1}{|\mathcal{I}|} \sum_{n\in \mathcal{I}} \hat{p}_j^{(n)}
                \end{gathered} \\[5pt]
			\end{equation*}
		\end{minipage} \\
		\hline 
		\\[-5pt]
		Estimated Relative Mean Squared Error (MSE) & \begin{minipage}{7cm}
			\begin{equation*} \label{eq:estimMSE}
				\underset{n\in \mathcal{I}}{\operatorname{med}}\left[\frac{\bigl(\hat{p}_j^{(n)} - p^*_j\bigr)^2 + \bigl(\hat{p}_j^{(n)} - \bar{p}_j\bigr)^2}{(p^*_j)^2}\right]\\[5pt]
			\end{equation*}
		\end{minipage} \\
		\hline 
		\\[-5pt]
		Estimated Coverage & \begin{minipage}{7.5cm}
            \begin{equation*} \label{eq:estimCov}
                \begin{gathered}
					\frac{|\mathcal{A}|}{|\mathcal{I}|} \\
                    \mathcal{A} = \Bigl\{\hat{p}_j \mid \bigl\lvert\hat{p}_j^{(n)} - p^*_j\bigr\rvert <  2 \sqrt{C_{j,j}^{(n)}}, n \in \mathcal{I} \Bigr\}\\[5pt]
                \end{gathered}
			\end{equation*}
		\end{minipage} \\
		\bottomrule
	\end{tabular}
	}
	\caption{Note that in the computation for relative forward simulation error the true initial condition is used. The sets of successful and failed runs are denoted $\mathcal{I}$ and $\mathcal{I}^c$ respectively. A run is deemed successful if the estimated parameters produce a relative coefficient error and relative forward simulation error that are both less than 2500\%. A run is also consider to have failed if \jlinl{NaN} values are produced during forward simulation. The metrics are computed over the set of successful runs. The operators $\operatorname{med}$ and $|\cdot|$ compute the median and the cardinality of a set respectively.
	}
	\label{table:metrics}
\end{table}

We say an algorithm failed if its estimated parameters cause forward simulation to return \jlinl{NaN} values or if either the relative forward simulation  error or relative coefficient  error is larger than 2500\%. Failed runs do not effect metrics averages which is to the benefit of the algorithm that has failed. 

\subsection{Corrupting the Data With Artificial Noise} \label{sec:artificialNoise}

For testing purposes, the data is corrupted with noise before it is passed as an input to the algorithms. When noise is additive Gaussian, we add noise proportional to the size the state variables as described in Equation \eqref{eq:nr}:
\begin{equation} \label{eq:nr}
	\begin{split}
		\noise_m\stackrel{iid}{\sim} \mathcal{N}(0, \Sigma_{nr})\\
		\Sigma = \operatorname{diag}\Biggl(\frac{\sigma_{nr} \|U\|_F^2}{M+1}\mathbf{1}_D\Biggr)
	\end{split}
\end{equation}
where $\sigma_{nr}$ is the noise ratio. For the multiplicative log-normal noise we use a scale matrix of $\sigma_{nr}\mathbb{I}_D$: 
\begin{equation} \label{eq:lognr}
	\begin{split}
		\log(\lognoise_m) \stackrel{iid}{\sim} \mathcal{N}(0, \sigma_{nr}\mathbb{I}_D).
	\end{split}
\end{equation}

\subsection{Motivating Example: Lorenz Oscillator} \label{sec:motiv}

The Lorenz Oscillator described by Equation \eqref{eq:lorenz} is a prototypical example of a chaotic system \citep{Sparrow1982}. Estimating the parameters in a chaotic system is difficult with output error methods because forward simulation is sensitive to the initial conditions given. In practice true (noiseless) initial conditions are not always available. The weak form methods do not suffer from this complication in the same way. 

\begin{figure}[!ht]
	\centering
	\includegraphics[width=0.27\textwidth]{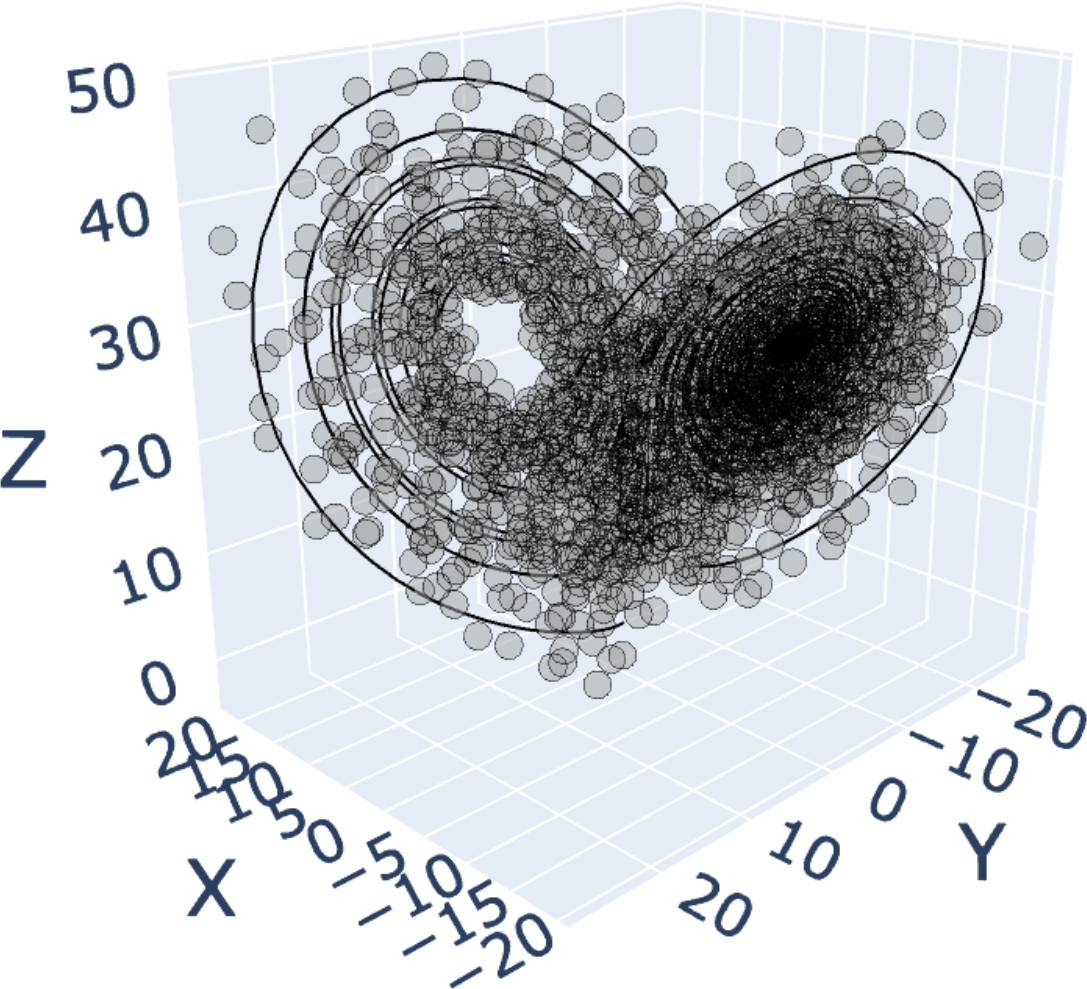} 
	\includegraphics[width=0.71\textwidth]{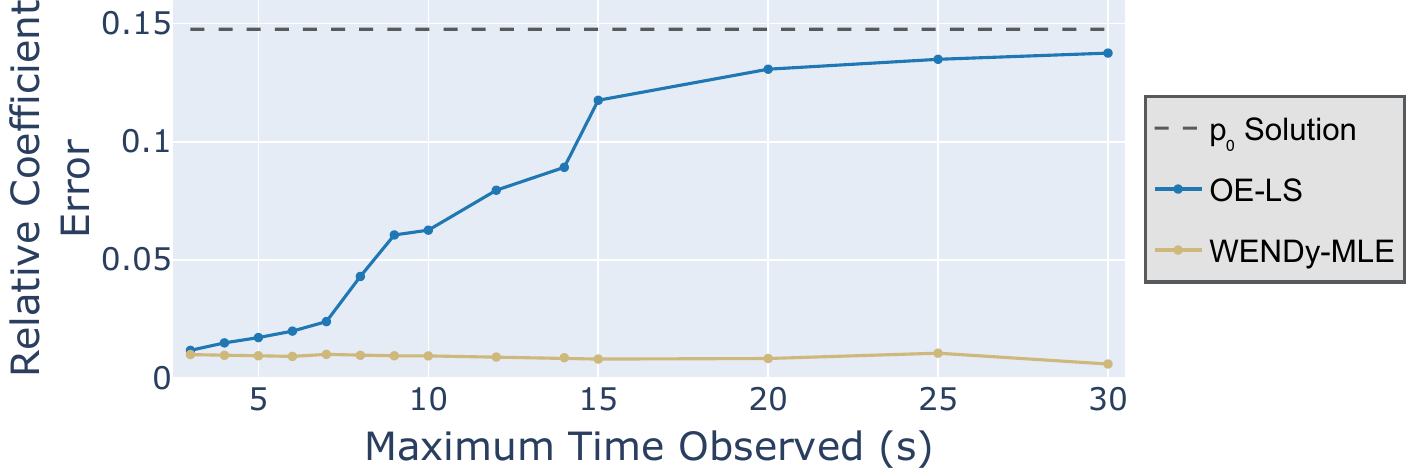}
	\caption{Left: the solution for the Lorenz oscillator in black and the corrupted data in grey dots. Right: the relative coefficient error for both the WENDy-MLE solver and the output error solver. }
	\label{fig:lorenz_experiment}
\end{figure}

To demonstrate this, we design an experiment where both solvers are given progressively more data from $[0, T]$ where $T \in \{3n\}_{n=1}^{10}$. One would hope that given more data, each algorithm would give progressively better estimates of the parameters, but in fact Fig.~\ref{fig:lorenz_experiment} shows that in fact the performance of OE-LS gets worse with more data.

For this experiment, we run each algorithm thirty times for each $T$. On each run, we vary additive Gaussian noise and initial guess for the parameters. Noise is added at a noise ratio of 10\% as described in Equation \eqref{eq:nr}. The data is given on a uniform grid with $\Delta T = 0.01$. The plot shows the mean coefficient relative error as described in Table~\ref{table:metrics} for both algorithms as well as the initial parameters. As more data becomes available, WENDy-MLE maintains an excellent estimate of the parameters while the OE-LS estimate degrades monotonically. 
The apparent contradiction of the OE-LS estimate degrading with more data is explained by the chaotic nature of the system, meaning that the objective cannot be evaluated accurately (due to floating point error accumulating in a catastrophic fashion) as well as significant non-convexity of the output loss. In contrast, WENDy-MLE does not require forwards simulation. Each element of the weak residual corresponds to a test function. This implicitly defines local regions where the effects of the parameters can be evaluated. This creates a smoother and better conditioned optimization problem.

\begin{table}[!ht]
	\centering
\begin{tabular}{crc}
	\toprule 
	 \textbf{Name} & \multicolumn{1}{c}{\textbf{ODE}} & \textbf{Parameters}  \\
	\midrule 
	\\[-5pt]
	Lorenz & \begin{minipage}{5.5cm} 
		\begin{equation} \label{eq:lorenz}
			\begin{aligned}
				\dot{u}_1 &= p_1(u_2-u_1) \\ 
				\dot{u}_2 &= u_1(p_2-u_3)-u_2 \\ 
				\dot{u}_3 &= u_1 u_2-p_3 u_3
			\end{aligned}
		\end{equation}
	\end{minipage} & 
	\begin{minipage}{5cm} \(\begin{aligned}
		t&\in[0,T] \text { where } T \in \{3n\}_{n=1}^{10}\\
		\ustate(0)&=[2, 1, 1]^T \\
		\trueParams&=[10, 28, 8/3]^T
	\end{aligned}\) \end{minipage}  \\\\[-5pt]
	\hline 
	\\[-5pt]
	Hindmarsh-Rose & \begin{minipage}{5.5cm} 
		\begin{equation} \label{eq:hindmarshRose}
			\begin{aligned}
			\dot{u}_1 &= p_1 u_2 - p_2 u_1^3 \\
			&+ p_3 u_1^2 - p_4 u_3 \\
			\dot{u}_2 &= p_5 -  p_6 u_1^2+ p_7 u_2 \\
			\dot{u}_3 &= p_8 u_1+  p_9 - p_{10} u_3
			\end{aligned}
		\end{equation}
	\end{minipage} & 
	\begin{minipage}{5cm} \(\begin{aligned}
		t&\in[0,10]\\
		\ustate(0)&=[-1.31,-7.6,-0.2]^T \\
		\trueParams&=[10,10,30,10,10,50, \\
		&\hspace{0.25cm} 10,0.04,0.0319,0.01]^T
	\end{aligned}\) \end{minipage} \\\\[-5pt]
	\hline 
	\\[-5pt]
	Goodwin & \begin{minipage}{5cm} 
		\begin{equation} \label{eq:goodwin3d}
			\begin{aligned}
			\dot{u}_1 &= \frac{p_1}{2.15 + p_3 u_3^{p_4}} - p_2  u_1 \\
			\dot{u}_2 &= p_5u_1- p_6u_2 \\
			\dot{u}_3 &= p_7u_2-p_8u_3
			\end{aligned}
		\end{equation}
	\end{minipage} & 
	\begin{minipage}{5.5cm}
		\(\begin{aligned}
			t &\in [0,80]\\
			\ustate(0) &= [0.3617, 0.9137, 1.3934]^T \\
			\trueParams &= [3.4884, 0.0969, 10, 0.0969, \\
			&\hspace{0.25cm}  0.0581, 0.0969, 0.0775]^T
		\end{aligned} \)
	\end{minipage} \\\\[-5pt]
	\hline 
	\\[-5pt]
	SIR-TDI & \begin{minipage}{5cm} 
		\begin{equation} \label{eq:sir} 
			\begin{aligned}
				\dot{u}_{1} &= -p_{1}  u_{1} + p_{3}  u_{2} \\
				&+ \frac{p_1 e^{-p_1  p_2}}{1 - e^{-p_1  p_2}} u_{3} \\
				\dot{u}_{2} &= p_{1}  u_{1} - p_{3}  u_{2} \\
				& - p_{4}  (1 - e^{-p_{5}  t^2})  u_{2} \\
				\dot{u}_{3} &= p_{4}  (1 - e^{-p_{5}  t^2})  u_{2} \\
				&- \frac{p_1 e^{-p_1  p_2}}{1 - e^{-p_1  p_2}}  u_{3}
			\end{aligned}
		\end{equation}
	\end{minipage} & 
	\begin{minipage}{5cm} 
		\(\begin{aligned}
			t &\in [0,50]\\
			\ustate(0) &= [1,0,0]^T \\
			\trueParams  &= [1.99, 1.5, 0.074, 0.113, \\
			& \hspace{0.25cm} 0.0024]^T 
		\end{aligned} \)
	\end{minipage} \\\\[-5pt]
	\bottomrule
\end{tabular}
\caption{Table of example systems that are presented in the results.}
\label{tab:odes}
\end{table}

\subsection{Test Suite} \label{sec:mega}

Moving beyond the Lorenz example, we run both the WENDy-MLE and OE-LS algorithm over the three systems of differential equations described in Table \ref{tab:odes}. For each system, we run the algorithm 100 times at varying noise ratios and subsamplings of the data $(M=256, 512, 1024)$. For a particular run, all algorithms are given the same data that has been corrupted by noise and the same initial guess for the parameters. The initial parameters are sampled from uniform priors specified in Table~\ref{tab:initParams}. These priors also defined the box constraints when they are employed. Higher levels of noise and fewer data points increases the difficulty of recovering the unknown parameters. The final time is fixed at the specified value of $T$, and the number of data points is controlled by adjusting $\Delta T$. In this section, plots show metrics for $M=256$, for plots of the metrics for $M=512$ and $M=1024$ see Figs \ref{fig:hindmarshFleshedOut}, \ref{fig:goodwinFleshedOut}, and \ref{fig:sirFleshedOut} in \ref{appendix:sup}.  

Hindmarsh Rose is linear in parameters and results were presented in the prior work of \citep{BortzMessengerDukic2023BullMathBiol}. The improved WENDy-MLE algorithm presented here slightly improves accuracy metrics on this example and adds significant robustness compared to the WENDy-IRLS. The Goodwin model is chosen because it is NiP. In particular, parameters of Hill functions are simultaneously identified which is of particular interest in many applications \cite{HIll1910JPhysiol,GoutelleMaurinRougierEtAl2008FundamemntalClinicalPharma}. The SIR-TDI model is NiP and was chosen to demonstrate WENDy-MLE's new ability to handle $\rhs$ with a direct dependence on $\t$. Summary information about each system is shown in Table \ref{tab:odes-desc}.

\begin{table}[!ht]
    \centering
	\begin{tabular}{lll}
		\toprule	
		\textbf{Name} & \textbf{Linear in Parameters} & \textbf{Noise Distribution} \\
		\midrule	
		Hindmarsh-Rose & Linear & Additive normal \\
		Goodwin (3D) & Nonlinear & Multiplicative log-normal  \\
		SIR-TDI & Nonlinear & Multiplicative log-normal   \\
		\bottomrule
	\end{tabular}
	\caption{Characteristics of the test problems.}
	\label{tab:odes-desc}
\end{table}

\subsubsection{Hindmarsh Rose} \label{sec:hindmarsh}
The Hindmarsh-Rose Model was initially developed to explain neuronal bursting \citep{HindmarshRose1984ProcRSocLondBBiolSci}. Both state variables $\ustate_1$ and $\ustate_2$ demonstrate this bursting. Estimation of the parameters is complicated by the differences in scales. In particular, $p^*_6 = -50$ and $p^*_9 = 0.0318$ which are separated by three orders of magnitude. The WENDy-MLE algorithm use of the likelihood addresses this concern.
\begin{figure}[H]
	\centering
    \includegraphics[width=1\textwidth]{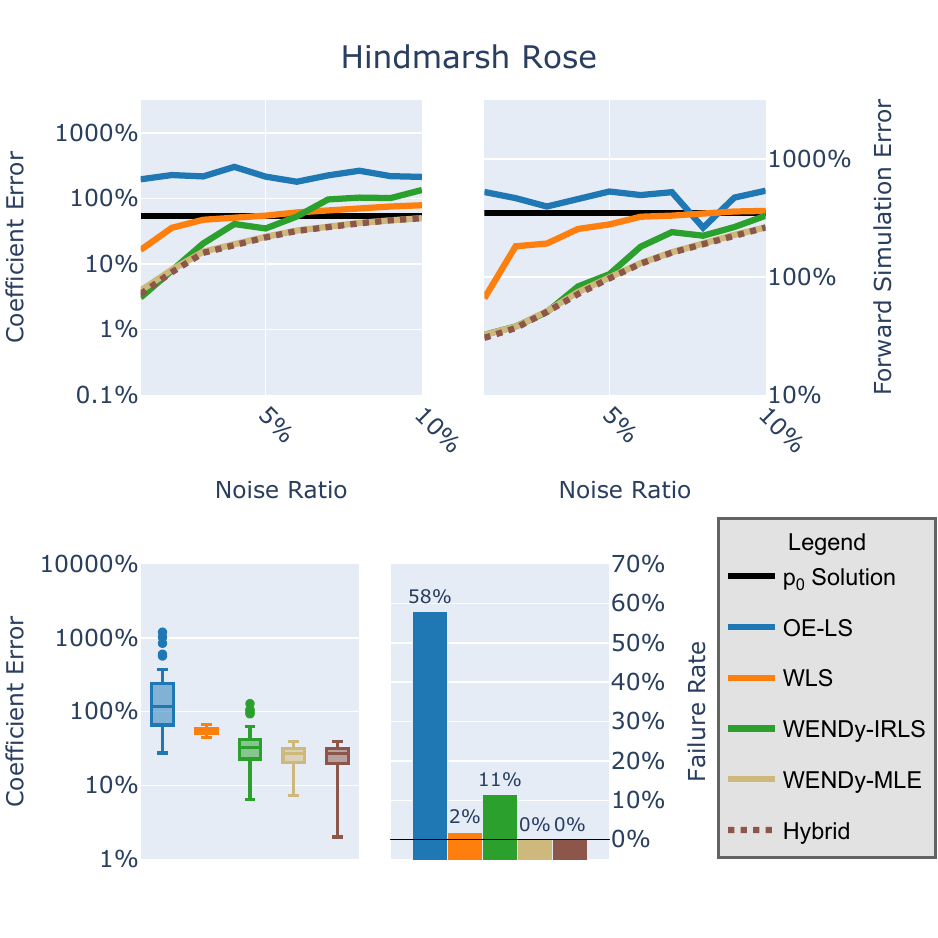}
	\caption{Metrics shown for OE-LS in blue, WLS in orange, WENDy-IRLS in green, WENDy-MLE in gold, the hybrid method in brown, and initial parameterization in black. Top left coefficient error vs noise ratio, top right forward simulation error vs noise ratio, bottom left box plot of algorithms coefficient relative error for all 100 runs at a noise ratio of 5\%, and bottom right average failure rate over all runs.}
	\label{fig:hindMega_accuracy}
\end{figure}

An initial motivation for this work was to improve performance on this example because WENDy-IRLS does not always converge, especially with less data and high noise. In contrast, Fig.~\ref{fig:hindMega_accuracy} demonstrates how the likelihood approach described here always converges as well as showing a mild improvement in accuracy in the cases where both methods converge.

Across all parameters, the median estimated relative squared bias, variance, and mean squared error are 13\%, 0.531\%, 16.6\% respectively. The median coverage across all parameters is 97\%. These metrics are shown in Fig.~\ref{fig:hindmarshRose_confInt} and Fig.~\ref{fig:hindmarshSup} in \ref{appendix:sup}.
For this example the WENDy-MLE algorithm has a bias that affects the MSE for all parameters. 
At all subsampling rates and for noise ratios less than 10\%, the bias stays below 25\% for all parameters except $p_8,p_9$ and $p_{10}$. These three parameters suffer from a lack of identifiability mentioned in \citep{BortzMessengerDukic2023BullMathBiol} which comes from adding noise proportional to the average norm of the state variable $\frac{\|\dstatemat\|_F^2}{M+1}$ rather than having noise be anisotropic across state components. 

Coverage at noise ratios below 5\% stays above 90\% for all variables. Uncertainty estimates for $p_8,p_9$ and $p_{10}$ stay sufficiently high resulting in a coverage >95\% for all noise levels in these parameters. On the other hand, the other parameters all have coverage rates monotonically decrease as the noise ratio increase. This is because the approximate likelihood given by Proposition \ref{prop:wres-dist} is poor when the noise is not sufficiently small. The linearization used to obtain the weak likelihood becomes a worse estimate for the true likelihood as the noise ratio increases. Also, when the data rate becomes sparser (smaller $M$) the coverage also degrades. Since the solution to the Hindmarsh-Rose system has discontinuities, having sufficient data around these events is paramount for any estimation method. 

\subsubsection{Goodwin} \label{sec:goodwin3d}
A simple example of a system of differential equations which is nonlinear in parameters is the Goodwin model which describes negative feedback control processes \citep{Calderhead2012PhD,Goodwin1965AdvancesinEnzymeRegulation}. This example was chosen because of the nonlinear right hand side. In particular, the right hand side contains a rational function known in the mathematical biology literature as a Hill function \cite{HIll1910JPhysiol,GoutelleMaurinRougierEtAl2008FundamemntalClinicalPharma}. 
All algorithms estimate the hill coefficient, $p_4$, in the exponent of the state variable. For $p^*_4=10$ sinusoidal oscillations occur in the state variables, while smaller values do not exhibit this behavior \citep{GonzeAbou-Jaoude2013PLoSONE}. 

\begin{figure}[!ht]
	\centering
	\includegraphics[width=1\textwidth]{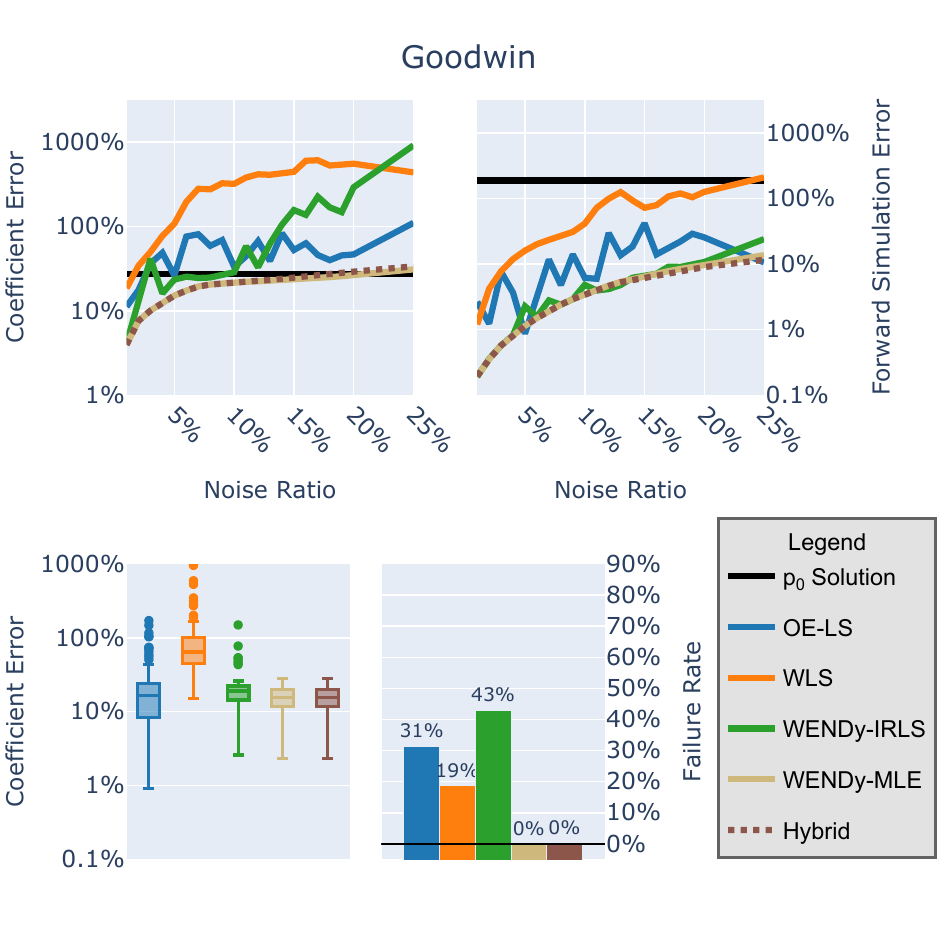}
	\caption{Metrics shown for OE-LS in blue, WLS in orange, WENDy-IRLS in green, WENDy-MLE in gold, the hybrid method in brown, and initial parameterization in black. Top left coefficient error vs noise ratio, top right forward simulation error vs noise ratio, bottom left box plot of algorithms coefficient relative error for all 100 runs at a noise ratio of 5\%, and bottom right average failure rate over all runs.}
	\label{fig:goodwin3dMega_accuracy}
\end{figure}

The results shown in Fig.~\ref{fig:goodwin3dMega_accuracy} show that both the WENDy-MLE and the hybrid algorithms perform reasonably well on this example while other algorithms have unreliable performance. Interestingly, WENDy-MLE barely outperforms the hybrid approach on coefficient error while the hybrid approach barely outperforms WENDy-MLE on forward simulation error. The failures of the OE-LS and the hybrid approach are primarily due to the coefficient error being above 2500\%. Sometimes these runs still have reasonable forward solve error and this accounts for the majority of the cases where the metrics from OE-LS do not match the hybrid approach.  

Because the log-normal noise distribution adds more nonlinearity to the weak likelihood, it is somewhat surprising that the WENDy-MLE algorithm performed this well. The linearization(s) made to derive the  likelihood led us to believe that it would be more limited in its effectiveness. It is also surprising that using WENDy-MLE as an initialization significantly improves on the performance of the forward based solver. Upon inspection, this is because the WENDy-MLE optimum is often within a domain of convergence for the forward based solver. 

Across all parameters, the median estimated relative squared bias, variance, and mean squared error are 0.0357\%, 0.388\%, 0.867\% respectively. The median coverage across all parameters is 100\%. The metrics for each parameter are shown in Fig.~\ref{fig:goodwin_confInt} and Fig.~\ref{fig:goodwinSup} 
in \ref{appendix:sup}. Bias drives the MSE for this example caused by the nonlinearity of $\rhs$ in both $\ustate$ and $\params$. In particular, $p_3$ exhibits the largest bias nearly reaching 100\%. This parameter struggles from a lack of identifiability. Because log-normal noise is used here, the magnitude of the noise is now proportional to each state $u_d$. Thus, the noise is anisotropic after the transformation, and this leads to the parameters connected to $u_3$ more identifiable. Coverage is particularly good for this example. Only the coverage for $p_1$ and $p_3$ ever falls below 95\%. 

\subsubsection{SIR-TDI} \label{sec:sir}
The susceptible-infected-recovered (SIR) model is pervasive in epidemiology. The system described in Equation \eqref{eq:sir} described an extension that allows for time delayed immunity (TDI) for parasitic deceases where there is a common source for infection \cite{ShonkwilerHerod2009}. 

\begin{figure}[!ht]
	\centering
	\includegraphics[width=1\textwidth]{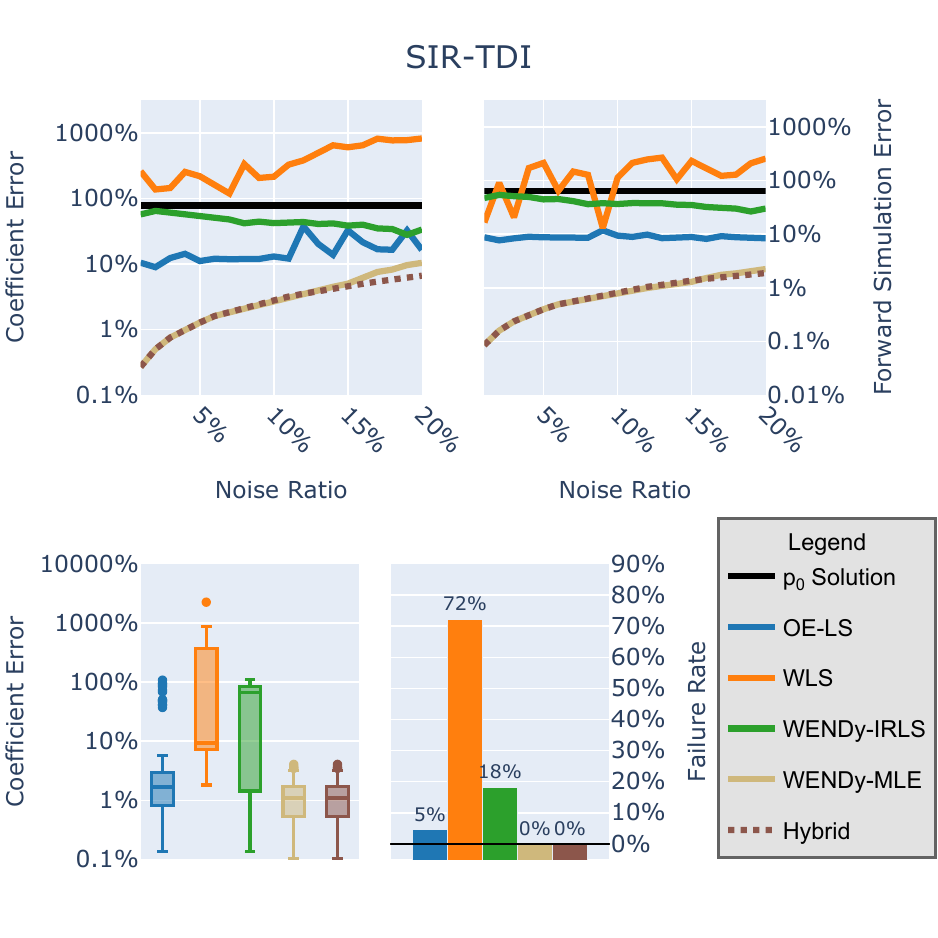}
	\caption{Metrics shown for OE-LS in blue, WENDy-MLE in gold, and initial parametrization in black. Top left coefficient error vs noise ratio, top right forward simulation error vs noise ratio, bottom left box plot of algorithms coefficient relative error for all 100 runs at a noise ratio of 5\%, and bottom right average failure rate over all runs.}
	\label{fig:sirMega_accuracy}
\end{figure}

The results are shown in Fig.~\ref{fig:sirMega_accuracy}. WENDy-MLE and hybrid method dominate the performance of other approaches. Also, unlike the other approaches, neither fails for any run thus demonstrating the larger domain of convergence for both approaches. It is noteworthy to mention that the WENDy-MLE algorithm was run with box constraints on this example. On the other example problems the constrained solver showed no performance improvements, but for this example WENDy-MLE's performance improved drastically with constraints. The constraints are quite loose and are the same as the those used in the uniform priors specified in Table~\ref{tab:initParams}. The improved performance of the constrained solver is most likely due to the non-linearity in the cost space caused by the highly nonlinear right hand side of Equation \eqref{eq:sir}. The nonlinearity is further exacerbated by multiplicative log-normal noise which requires applying logarithms on both sides of the equation as described in Section \ref{sec:log}.

Across all parameters, the median estimated relative squared bias, variance, and mean squared error are 0.00418\%, 0.0681\%, 0.165\% respectively. The median coverage across all parameters is 100\%. These metrics are shown in Fig~\ref{fig:sir_confInt} and Fig.~\ref{fig:sirSup}
in \ref{appendix:sup}. Bias, variance and MSE are all below 5\% for all parameters and subsamplings (values of $M$) except for $p_5$ and $M=256$ at noise ratios larger than 15\%. $p_5$ is a nonlinear parameter that is an exponential rate. Its true value of $0.0024$ is relatively small compared to the other parameters in this system. With less data and higher noise, it is not surprising that this parameter becomes harder to identify.
The uncertainty in WENDy-MLE estimator is overestimated resulting in a coverage larger than the nominal 95\% for all parameters across the board. This higher-than-nominal coverage is preferable to the alternative.

\section{Discussion}\label{sec:discussion}

The WENDy framework presented here addresses a more general class of problems than the previous weak form methods. In particular, WENDy-MLE can handle systems of ODEs which are nonlinear in the parameters. Furthermore, our method can leverage the distribution for both cases of additive Gaussian Noise and multiplicative log-normal noise. Overall, the WENDy-MLE algorithm estimates parameters more accurately and fails less often than both previous weak form methods as well as output error methods. This is due to the implicit regularization that the $\operatorname{logdet}$ term of the weak likelihood in Equation \eqref{eq:wnll}. This term penalizes the parameters that increase the eigenvalues of the covariance $\wCov$.
We summarize some qualitative takeaways from our experiments in Table~\ref{table:summary}. 

\pagebreak 
\renewcommand*\arraystretch{1.75}
\begin{table}[!ht]
    \resizebox{\textwidth}{!}{
    \begin{tabular}{cccccc}
        \toprule
        &  &   & \multicolumn{3}{c}{\textbf{Noise}}  \\
        \cline{4-6}
         \multirow{-2}{*}{\textbf{Algorithm}} &
        \multirow{-2}{*}{\textbf{Chaotic}} & \multirow{-2}{*}{\textbf{Nonlinear In Parameters}} & Noiseless &   Gaussian &  LogNormal \\
        \midrule
        OE-LS & \setcounter{footnote}{4}\xmark\footnotemark & \checkmark & \checkmark & \checkmark & \checkmark\\       
        
        WLS & \checkmark & \setcounter{footnote}{5}\xmark\footnotemark  & \checkmark\checkmark & \xmark  & \setcounter{footnote}{5}\xmark\footnotemark \\       
        
        WENDy-IRLS & \checkmark\checkmark & \setcounter{footnote}{5}\xmark\footnotemark  & \setcounter{footnote}{6}\checkmark\footnotemark & \checkmark & \setcounter{footnote}{5}\xmark\footnotemark \\       
        
        WENDy-MLE & \checkmark\checkmark  & \setcounter{footnote}{7}\checkmark\checkmark\footnotemark & \setcounter{footnote}{6} \checkmark \footnotemark & \checkmark\checkmark  & \checkmark\checkmark \\  
        \bottomrule
    \end{tabular}
    }
	\caption{Here we provide high-level advice on the performance of the discussed methods applied to different classes of problems. We compare OE-LS, WLS, WENDy-IRLS, and  WENDy-MLE. A double check \checkmark\checkmark\, indicates the algorithm excels under most circumstances in our testing. A single \checkmark\, suggests that the algorithm can work, but requires oversight due to a reduced domain of convergence for initial parameter estimates as well as other numerical discretization challenges. An \xmark\, indicates that the algorithm is not suited to the problem (e.g., WLS and WENDy-IRLS only work for linear-in-parameter ODEs and without correction, noisy data will incur bias for WLS).}\label{table:summary}
\end{table}

\setcounter{footnote}{5}
\footnotetext{OE-LS can work on short time horizons but in general should not be used for chaotic systems.}
\setcounter{footnote}{6}
\footnotetext{In the Julia implementation, both of these algorithms have been extended by replacing the least square problems with a nonlinear least squares problems. Similarly, the infrastructure makes it possible to run these algorithms with log-normal noise as well.}
\setcounter{footnote}{7}
\footnotetext{Because WENDy-IRLS and WENDy-MLE assume measurement noise is present, computation of the covariance is ill-posed for noiseless data. Our software implementation automatically switches to WLS in this case. In some instance we have seen that incorporating covariance information to estimate parameters for noiseless data can be superior to WLS. We conjecture the algorithms may be correcting for the numerical error in the solver that generated the artificial data, but this remains a topic for future research.}
\setcounter{footnote}{8}
\footnotetext{For the vast majority of the examples, the domain of convergence is substantially larger than for OE-LS. However, as with all iterative methods, WENDy-MLE benefits from wise initial parameter estimates.}

There are known limitations to our approach. First, WENDy-MLE is based on maximizing an approximation of the likelihood of the weak residual. Accordingly, the form of the ODE must be such that integration-by-parts is possible.  We also assume that the measurement noise is either additive Gaussian noise or multiplicative log-normal noise. We also require that all state variables are observed, which may not be true in some applications. Another limitation is that we assume observational data is provided on a uniform grid, though in principle, there is nothing inhibiting extension to the non-uniform case with a different quadrature rule. Finally, like almost all other methods that do general parameter estimation, we face a non-convex optimization problem for which we cannot guarantee finding the global minimizer for arbitrary initialization. Thus, in practice it is often necessary to initialize multiple times for best results.

\subsection{Robustness in the Presence of Noise}
Like other weak form methods, the true strength of the approach lies in the ability to perform well in the presence of noise\footnote{Smoothing can help (up to a point), see \cite{MessengerBortz2024IMAJNumerAnal}.}. No pre-processing or smoothing of noisy input data is necessary for our algorithm to be successful. Instead, the set of test functions that are chosen when forming $\dTestFun$ and $\dot{\dTestFun}$ are built adaptively from the data as described in \ref{appendix:TestFuns}. Furthermore, relying on the intuition from theoretical results, the discrete weak form residual is analogous to considering the continuous weak form solution. This means that our method is relying on a more relaxed topology, and this leads to algorithms with a cost space that is smoother and more easily navigable. 

\subsection{Bias} 

When $\rhs$ is linear with respect to the state variable $\ustate$ the WENDy estimator will be unbiased. When this is not the case the Gaussian approximation leads to a bias that increases in magnitude with the noise. In other words, the approximation provided by Proposition \ref{prop:wres-dist} is only valid when the noise is sufficiently small. The severity of nonlinearity of $\rhs$ with respect to the state variable $\ustate$ determines the noise regime where Proposition \ref{prop:wres-dist} holds. This is not to say that output error methods do not also suffer from bias. In general, there is no guarantee on the consistency of OE-LS.

\begin{figure}[!ht]
    \centering
    \includegraphics[width=0.3\textwidth]{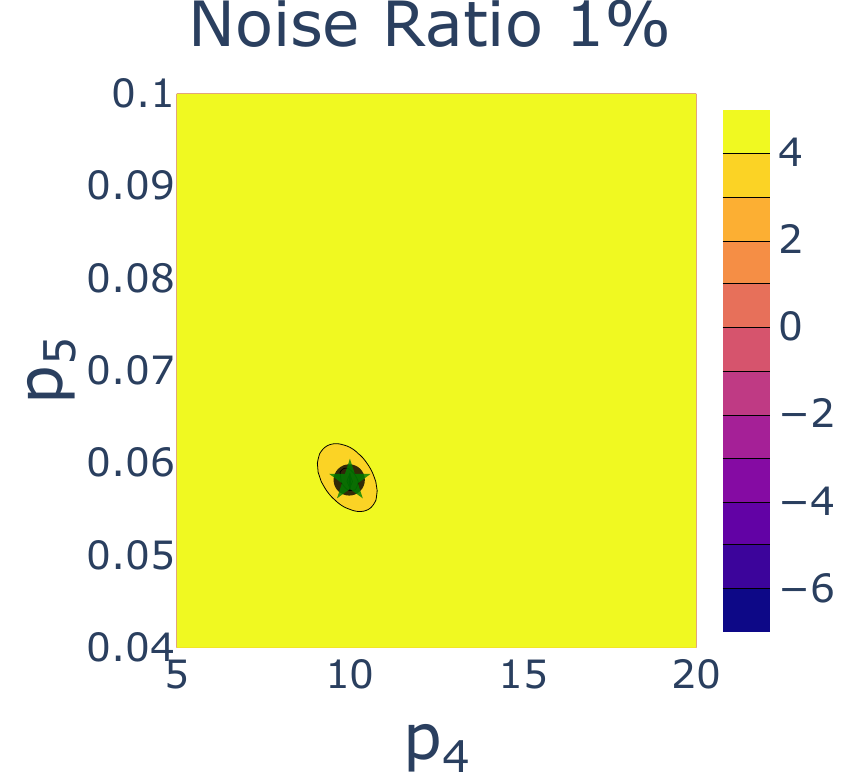}
    \includegraphics[width=0.3\textwidth]{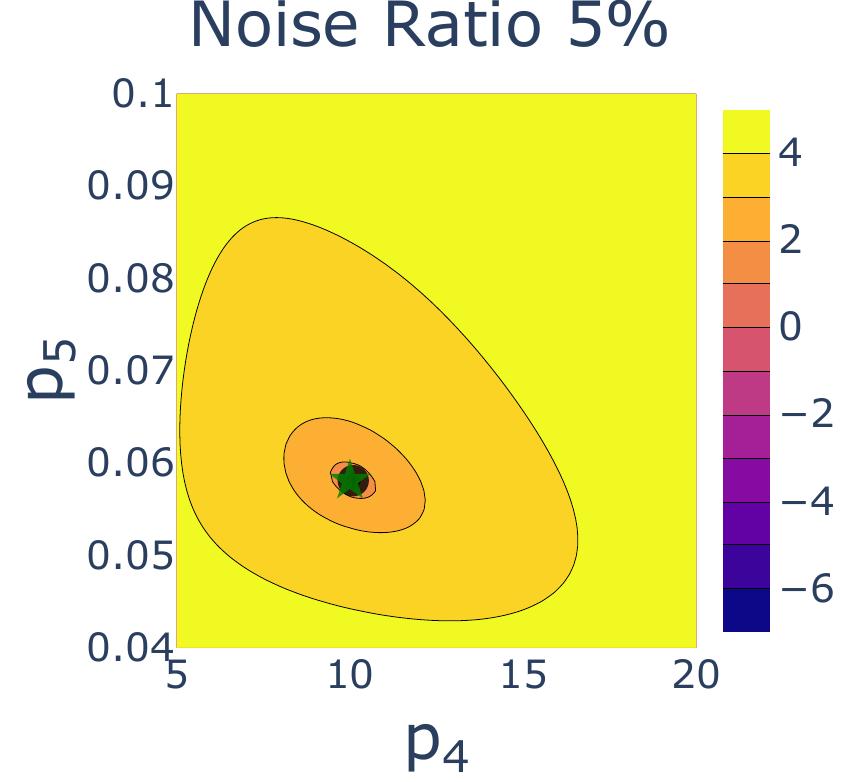}
    \includegraphics[width=0.3\textwidth]{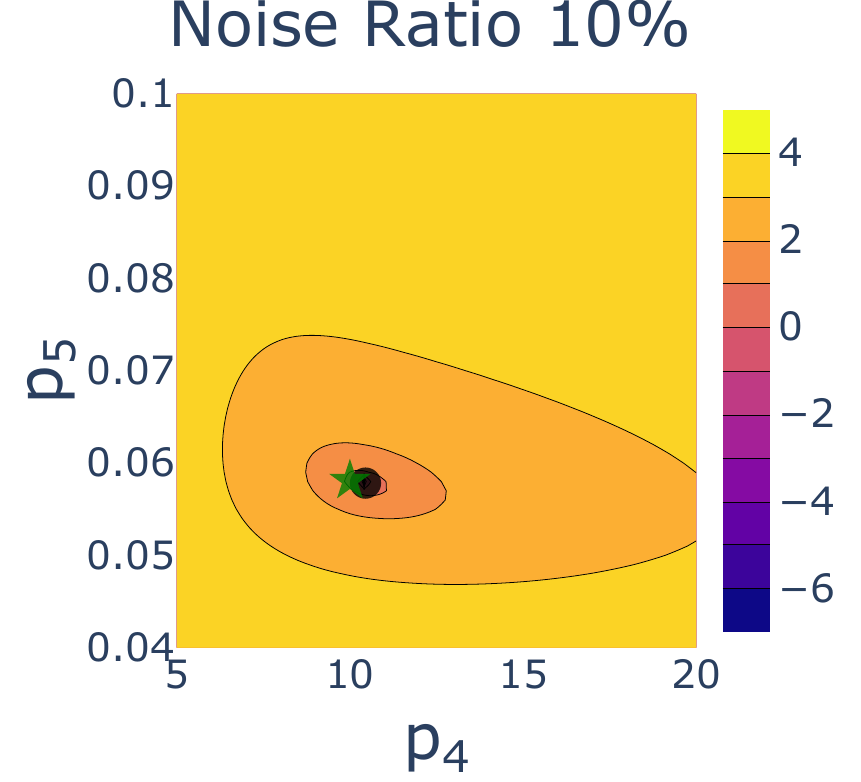}
    \includegraphics[width=0.3\textwidth]{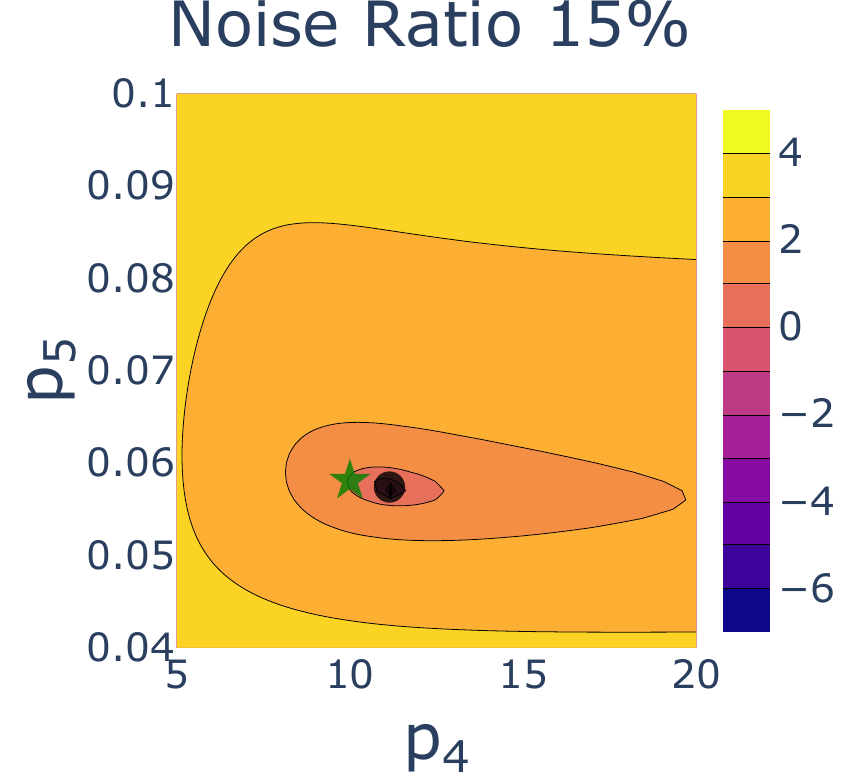}
    \includegraphics[width=0.3\textwidth]{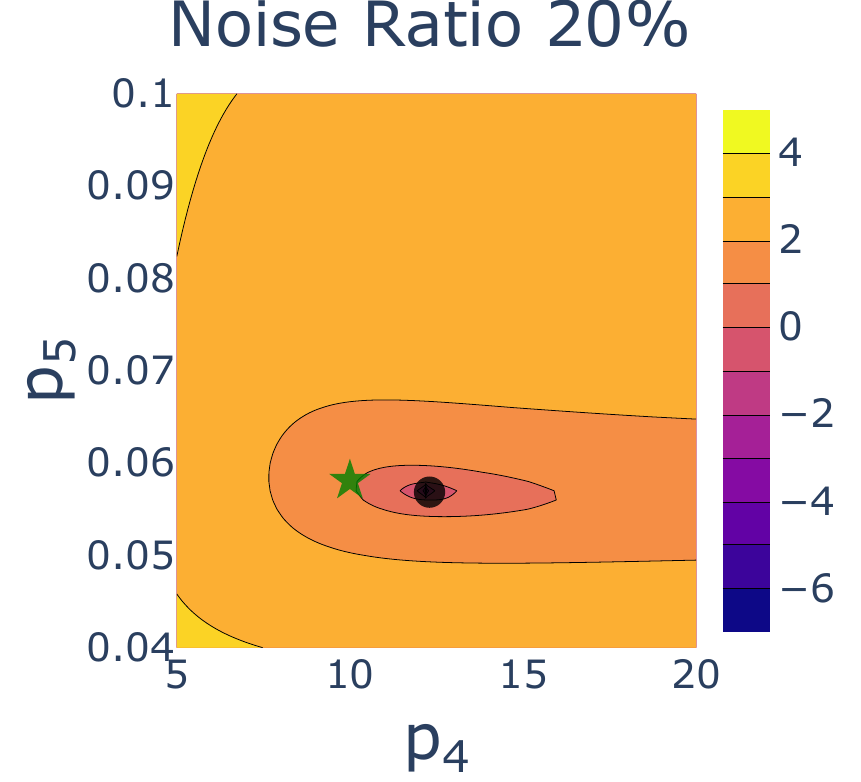}
    \includegraphics[width=0.3\textwidth]{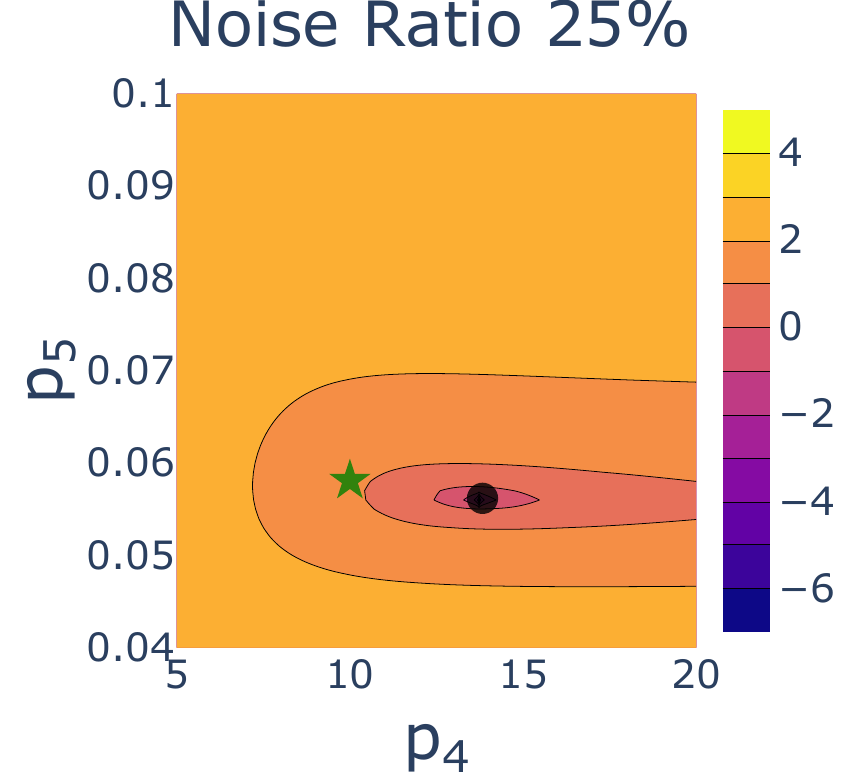}
    \caption{The WENDy algorithm is run on the Goodwin problem with all parameters fixed to truth except $p_4$ and $p_5$ for a variety of noise levels. The truth is shown as a green star, and the WENDy optimum is identified with a black circle. One can notice that the optimum moves further from truth as noise increases.}
    \label{fig:bias}
\end{figure}

The Goodwin system shown in Equation \eqref{eq:goodwin3d} is run with all other parameters fixed to their true values except for $p_4$ and $p_5$. At each noise level we sample 100 different instances of noise to corrupt the true data. Then we build the weak likelihood function for each instance of noise. The week likelihood for each instance of noise are evaluated on the grid and are then averaged over the instances of noise to produce the contour plots shown in Fig.~\ref{fig:bias}. As the noise ratio is increased, the local optimum is shifted away from truth. This is caused by the higher order terms that are neglected in the linearization used by  Proposition \ref{prop:wres-dist}. As the noise becomes larger, the nonlinear terms cause a bias in the maximum likelihood estimate.

\subsection{Computational Cost} 

A strength of the weak form methods is that no forward simulation is necessary at each step of the optimization routine. When the system of differential equations is small and not stiff, then standard forward solve methods may be more efficient and the main benefit to using a weak form method like ours is that of robustness to both noise and to poor initialization. However, as the dimensionality of state variables and stiffness of the problem increases, the forward solves become more computationally expensive, whereas weak form methods are unaffected. Instead the cost of the weak form methods dominantly scale with the number of test functions. See \ref{appendix:compCost} for an analysis of computational cost of weak likelihood and its first and second order derivatives.

\section{Conclusion} \label{sec:conclusion}

In this work, the WENDy algorithm was extended to a more general case of differential equations where the right hand side may be nonlinear in parameters. Furthermore, the approach explicitly supports both additive Gaussian and multiplicative log-normal noise. The likelihood based approach improved on the convergence and accuracy compared to the previous work. Also, the new Julia implementation improved the usability and efficiency of the algorithm. 

The likelihood derived is limited by the linearization used. Further investigation of higher order expansions or other derivation strategies are necessary to handle high nonlinearities and noise levels. The computational advantage of the approach should be more apparent in the extension of the algorithm to larger systems of ordinary differential equations and partial differential equations. Also, further optimization of test function selection to recover maximum information in the weak form is another area of further exploration.

\section{Acknowledgments} \label{sec:ack}

The authors would like to thank  April Tran, Jack Krebsbach, and Will Houser (University of Colorado) for insights regarding software development for weak form methods.

This work is supported in part by the National Institute of General Medical Sciences grant R35GM149335, National Science Foundation grant 
2109774, National Institute of Food and Agriculture grant 2019-67014-29919, and by the Department of Energy, Office of Science, Advanced Scientific Computing Research under Award Number DE-SC0023346.

\section*{Conflict of Interest}
The authors declare that they have no conflict of interest.
\renewcommand{\newblock}{}
\bibliographystyle{abbrv}
\bibliography{NonlinearWENDy4ODE}

\begin{thebibliography}{10}

\bibitem{AbbottAluthgeN3N5EtAl2023}
M.~Abbott, D.~Aluthge, N3N5, V.~Puri, C.~Elrod, S.~Schaub, C.~Lucibello, J.~Bhattacharya, J.~Chen, K.~Carlsson, and M.~Gelbrecht.
\newblock Mcabbott/{{Tullio}}.jl: V0.3.7.
\newblock Zenodo, Oct. 2023.

\bibitem{AldoghaitherLiuLaleg-Kirati2015SIAMJSciComput}
A.~Aldoghaither, D.-Y. Liu, and T.-M. {Laleg-Kirati}.
\newblock Modulating {{Functions Based Algorithm}} for the {{Estimation}} of the {{Coefficients}} and {{Differentiation Order}} for a {{Space-Fractional Advection-Dispersion Equation}}.
\newblock {\em SIAM J. Sci. Comput.}, 37(6):A2813--A2839, Jan. 2015.

\bibitem{Atkinson1989}
K.~Atkinson.
\newblock {\em An {{Introduction}} to {{Numerical Analysis}}}.
\newblock Wiley, New York, NY, 2nd edition, 1989.

\bibitem{BezansonEdelmanKarpinskiEtAl2017SIAMRev}
J.~Bezanson, A.~Edelman, S.~Karpinski, and V.~B. Shah.
\newblock Julia: {{A Fresh Approach}} to {{Numerical Computing}}.
\newblock {\em SIAM Rev.}, 59(1):65--98, Jan. 2017.

\bibitem{BortzMessengerDukic2023BullMathBiol}
D.~M. Bortz, D.~A. Messenger, and V.~Dukic.
\newblock Direct {{Estimation}} of {{Parameters}} in {{ODE Models Using WENDy}}: {{Weak-form Estimation}} of {{Nonlinear Dynamics}}.
\newblock {\em Bull. Math. Biol.}, 85(110), 2023.

\bibitem{BortzMessengerTran2024NumericalAnalysisMeetsMachineLearning}
D.~M. Bortz, D.~A. Messenger, and A.~Tran.
\newblock Weak form-based data-driven modeling: {{Computationally Efficient}} and {{Noise Robust Equation Learning}} and {{Parameter Inference}}.
\newblock In S.~Mishra and A.~Townsend, editors, {\em Numerical {{Analysis Meets Machine Learning}}}, volume~25 of {\em Handbook of {{Numerical Analysis}}}, pages 54--82. Elsevier, 2024.

\bibitem{BrunelClairondAlche-Buc2014JAmStatAssoc}
N.~J.-B. Brunel, Q.~Clairon, and F.~{d'Alch{\'e}-Buc}.
\newblock Parametric {{Estimation}} of {{Ordinary Differential Equations With Orthogonality Conditions}}.
\newblock {\em J. Am. Stat. Assoc.}, 109(505):173--185, Jan. 2014.

\bibitem{BruntonProctorKutz2016ProcNatlAcadSci}
S.~L. Brunton, J.~L. Proctor, and J.~N. Kutz.
\newblock Discovering governing equations from data by sparse identification of nonlinear dynamical systems.
\newblock {\em Proc. Natl. Acad. Sci.}, 113(15):3932--3937, Apr. 2016.

\bibitem{Calderhead2012PhD}
B.~Calderhead.
\newblock {\em Differential {{Geometric MCMC Methods}} and {{Applications}}}.
\newblock PhD thesis, University of Glasgow, 2012.

\bibitem{Chesher1991Biometrika}
A.~Chesher.
\newblock The effect of measurement error.
\newblock {\em Biometrika}, 78(3):451--462, 1991.

\bibitem{ConradMarzoukPillaiEtAl2016JAmStatAssoc}
P.~R. Conrad, Y.~M. Marzouk, N.~S. Pillai, and A.~Smith.
\newblock Accelerating {{Asymptotically Exact MCMC}} for {{Computationally Intensive Models}} via {{Local Approximations}}.
\newblock {\em J. Am. Stat. Assoc.}, 111(516):1591--1607, Oct. 2016.

\bibitem{DukicLopesPolson2012JAmStatAssoc}
V.~Dukic, H.~F. Lopes, and N.~G. Polson.
\newblock Tracking {{Epidemics With Google Flu Trends Data}} and a {{State-Space SEIR Model}}.
\newblock {\em J. Am. Stat. Assoc.}, 107(500):1410--1426, Dec. 2012.

\bibitem{Durbin1954RevIntStatInst}
J.~Durbin.
\newblock Errors in {{Variables}}.
\newblock {\em Rev. Int. Stat. Inst.}, 22(1/3):23, 1954.

\bibitem{DussaultMigotOrban2024MathProgram}
J.-P. Dussault, T.~Migot, and D.~Orban.
\newblock Scalable adaptive cubic regularization methods.
\newblock {\em Math. Program.}, 207(1-2):191--225, Sept. 2024.

\bibitem{GonzeAbou-Jaoude2013PLoSONE}
D.~Gonze and W.~{Abou-Jaoud{\'e}}.
\newblock The {{Goodwin Model}}: {{Behind}} the {{Hill Function}}.
\newblock {\em PLoS ONE}, 8(8):e69573, Aug. 2013.

\bibitem{Goodwin1965AdvancesinEnzymeRegulation}
B.~C. Goodwin.
\newblock Oscillatory behavior in enzymatic control processes.
\newblock {\em Advances in Enzyme Regulation}, 3:425--437, Jan. 1965.

\bibitem{GoutelleMaurinRougierEtAl2008FundamemntalClinicalPharma}
S.~Goutelle, M.~Maurin, F.~Rougier, X.~Barbaut, L.~Bourguignon, M.~Ducher, and P.~Maire.
\newblock The {{Hill}} equation: A review of its capabilities in pharmacological modelling.
\newblock {\em Fundamemntal Clinical Pharma}, 22(6):633--648, Dec. 2008.

\bibitem{GowdaMaCheliEtAl2021ACMCommunComputAlgebra}
S.~Gowda, Y.~Ma, A.~Cheli, M.~Gw{\'o}{\'z}zd{\'z}, V.~B. Shah, A.~Edelman, and C.~Rackauckas.
\newblock High-performance symbolic-numerics via multiple dispatch.
\newblock {\em ACM Commun. Comput. Algebra}, 55(3):92--96, Sept. 2021.

\bibitem{Greenberg1951NACATN2340}
H.~Greenberg.
\newblock A survey of methods for determining stability parameters of an airplane from dyanmics flight measurements.
\newblock Technical Report NACA TN 2340, Ames Aeronautical Laboratory, Moffett Field, CA, Apr. 1951.

\bibitem{GurevichReinboldGrigoriev2019Chaos}
D.~R. Gurevich, P.~A.~K. Reinbold, and R.~O. Grigoriev.
\newblock Robust and optimal sparse regression for nonlinear {{PDE}} models.
\newblock {\em Chaos}, 29(10):103113, Oct. 2019.

\bibitem{HallMa2014JRStatSocB}
P.~Hall and Y.~Ma.
\newblock Quick and easy one-step parameter estimation in differential equations.
\newblock {\em J. R. Stat. Soc. B}, 76(4):735--748, Sept. 2014.

\bibitem{HIll1910JPhysiol}
A.~V. Hill.
\newblock The possible effects of the aggregation of the molecules of haemoglobin on its dissociation curves.
\newblock {\em J. Physiol.}, 40(suppl):iv--vii, Dec. 1910.

\bibitem{HindmarshRose1984ProcRSocLondBBiolSci}
J.~L. Hindmarsh and R.~M. Rose.
\newblock A model of neuronal bursting using three coupled first order differential equations.
\newblock {\em Proc R Soc Lond B Biol Sci}, 221(1222):87--102, Mar. 1984.

\bibitem{Janiczek2010BullPolAcadSciTechSci}
T.~Janiczek.
\newblock Generalization of the modulating functions method into the fractional differential equations.
\newblock {\em Bull. Pol. Acad. Sci. Tech. Sci.}, 58(4), Jan. 2010.

\bibitem{JouffroyReger20152015IEEEConfControlApplCCA}
J.~Jouffroy and J.~Reger.
\newblock Finite-time simultaneous parameter and state estimation using modulating functions.
\newblock In {\em 2015 {{IEEE Conf}}. {{Control Appl}}. {{CCA}}}, pages 394--399, Sydney, Australia, Sept. 2015. IEEE.

\bibitem{KMogensenNRiseth2018JOSS}
P.~K~Mogensen and A.~N~Riseth.
\newblock Optim: {{A}} mathematical optimization package for {{Julia}}.
\newblock {\em JOSS}, 3(24):615, Apr. 2018.

\bibitem{KennedyDukicDwyer2014AmNat}
D.~A. Kennedy, V.~Dukic, and G.~Dwyer.
\newblock Pathogen {{Growth}} in {{Insect Hosts}}: {{Inferring}} the {{Importance}} of {{Different Mechanisms Using Stochastic Models}} and {{Response-Time Data}}.
\newblock {\em Am. Nat.}, 184(3):407--423, Sept. 2014.

\bibitem{KoldaBader2009SIAMRev}
T.~G. Kolda and B.~W. Bader.
\newblock Tensor {{Decompositions}} and {{Applications}}.
\newblock {\em SIAM Rev.}, 51(3):455--500, Aug. 2009.

\bibitem{LiuChangChen2016NonlinearDyn}
C.-S. Liu, J.-R. Chang, and Y.-W. Chen.
\newblock The recovery of external force in nonlinear system by using a weak-form integral method.
\newblock {\em Nonlinear Dyn.}, 86(2):987--998, Oct. 2016.

\bibitem{LiuLaleg-Kirati2015SignalProcessing}
D.-Y. Liu and T.-M. {Laleg-Kirati}.
\newblock Robust fractional order differentiators using generalized modulating functions method.
\newblock {\em Signal Processing}, 107:395--406, Feb. 2015.

\bibitem{Ljung1999}
L.~Ljung.
\newblock {\em System Identification: Theory for the User}.
\newblock Prentice {{Hall}} Information and System Sciences Series. Prentice Hall PTR, Upper Saddle River, NJ, 2nd edition, 1999.

\bibitem{Ljung2017WileyEncyclopediaofElectricalandElectronicsEngineering}
L.~Ljung.
\newblock {\em System {{Identification}}}, pages 1--19.
\newblock John Wiley \& Sons, Inc., Hoboken, NJ, USA, May 2017.

\bibitem{LoebCahen1965IEEETransAutomControl}
J.~M. Loeb and G.~M. Cahen.
\newblock More about process identification.
\newblock {\em IEEE Trans. Autom. Control}, 10(3):359--361, July 1965.

\bibitem{McGoffMukherjeePillai2015StatSurv}
K.~McGoff, S.~Mukherjee, and N.~Pillai.
\newblock Statistical inference for dynamical systems: {{A}} review.
\newblock {\em Stat. Surv.}, 9:209--252, Jan. 2015.

\bibitem{MessengerBortz2021JComputPhys}
D.~A. Messenger and D.~M. Bortz.
\newblock Weak {{SINDy For Partial Differential Equations}}.
\newblock {\em J. Comput. Phys.}, 443:110525, Oct. 2021.

\bibitem{MessengerBortz2021MultiscaleModelSimul}
D.~A. Messenger and D.~M. Bortz.
\newblock Weak {{SINDy}}: {{Galerkin-Based Data-Driven Model Selection}}.
\newblock {\em Multiscale Model. Simul.}, 19(3):1474--1497, 2021.

\bibitem{MessengerBortz2022PhysicaD}
D.~A. Messenger and D.~M. Bortz.
\newblock Learning mean-field equations from particle data using {{WSINDy}}.
\newblock {\em Physica D}, 439:133406, Nov. 2022.

\bibitem{MessengerBortz2024IMAJNumerAnal}
D.~A. Messenger and D.~M. Bortz.
\newblock Asymptotic consistency of the {{WSINDy}} algorithm in the limit of continuum data.
\newblock {\em IMA J. Numer. Anal.}, page drae086, Dec. 2024.

\bibitem{MessengerBurbyBortz2024SciRep}
D.~A. Messenger, J.~W. Burby, and D.~M. Bortz.
\newblock Coarse-{{Graining Hamiltonian Systems Using WSINDy}}.
\newblock {\em Sci. Rep.}, 14(14457):1--24, June 2024.

\bibitem{MessengerDallAneseBortz2022ProcThirdMathSciMachLearnConf}
D.~A. Messenger, E.~Dall'Anese, and D.~M. Bortz.
\newblock Online {{Weak-form Sparse Identification}} of {{Partial Differential Equations}}.
\newblock In {\em Proc. {{Third Math}}. {{Sci}}. {{Mach}}. {{Learn}}. {{Conf}}.}, volume 190 of {\em Proceedings of {{Machine Learning Research}}}, pages 241--256. PMLR, 2022.

\bibitem{MessengerTranDukicEtAl2024SIAMNews}
D.~A. Messenger, A.~Tran, V.~Dukic, and D.~M. Bortz.
\newblock The {{Weak Form Is Stronger Than You Think}}.
\newblock {\em SIAM News}, 57(8), Oct. 2024.

\bibitem{MessengerWheelerLiuEtAl2022JRSocInterface}
D.~A. Messenger, G.~E. Wheeler, X.~Liu, and D.~M. Bortz.
\newblock Learning {{Anisotropic Interaction Rules}} from {{Individual Trajectories}} in a {{Heterogeneous Cellular Population}}.
\newblock {\em J. R. Soc. Interface}, 19(195):20220412, Oct. 2022.

\bibitem{MigotOrbanSoaresSiqueira2024}
T.~Migot, D.~Orban, and A.~Soares~Siqueira.
\newblock {{JSOSolvers}}.jl: {{JuliaSmoothOptimizers}} optimization solvers.
\newblock Zenodo, Oct. 2024.

\bibitem{NardiniBortz2019InverseProbl}
J.~T. Nardini and D.~M. Bortz.
\newblock The influence of numerical error on parameter estimation and uncertainty quantification for advective {{PDE}} models.
\newblock {\em Inverse Probl.}, 35(6):065003, May 2019.

\bibitem{PalHoltorfLarssonEtAl2024arXiv240316341}
A.~Pal, F.~Holtorf, A.~Larsson, T.~Loman, Utkarsh, F.~Sch{\"a}efer, Q.~Qu, A.~Edelman, and C.~Rackauckas.
\newblock {{NonlinearSolve}}.jl: {{High-Performance}} and {{Robust Solvers}} for {{Systems}} of {{Nonlinear Equations}} in {{Julia}}.
\newblock {\em arXiv:2403.16341}, Mar. 2024.

\bibitem{PantazisTsamardinos2019Bioinformatics}
Y.~Pantazis and I.~Tsamardinos.
\newblock A unified approach for sparse dynamical system inference from temporal measurements.
\newblock {\em Bioinformatics}, 35(18):3387--3396, Sept. 2019.

\bibitem{PatraUnbehauen1995IntJControl}
A.~Patra and H.~Unbehauen.
\newblock Identification of a class of nonlinear continuous-time systems using {{Hartley}} modulating functions.
\newblock {\em Int. J. Control}, 62(6):1431--1451, Dec. 1995.

\bibitem{PearsonLee1985Control-TheoryAdvTechnol}
A.~E. Pearson and F.~C. Lee.
\newblock Parameter identification of linear differential systems via {{Fourier}} based modulating functions.
\newblock {\em Control-Theory Adv. Technol.}, 1(4):239--266, Dec. 1985.

\bibitem{PinAssaloneLoveraEtAl2015IEEETransAutomatContr}
G.~Pin, A.~Assalone, M.~Lovera, and T.~Parisini.
\newblock Non-{{Asymptotic Kernel-Based Parametric Estimation}} of {{Continuous-time}}.
\newblock {\em IEEE Trans. Automat. Contr.}, pages 1--1, 2015.

\bibitem{PinChenParisini2017Automatica}
G.~Pin, B.~Chen, and T.~Parisini.
\newblock Robust finite-time estimation of biased sinusoidal signals: {{A}} volterra operators approach.
\newblock {\em Automatica}, 77:120--132, Mar. 2017.

\bibitem{RackauckasNie2017JORS}
C.~Rackauckas and Q.~Nie.
\newblock {{DifferentialEquations}}.jl -- {{A Performant}} and {{Feature-Rich Ecosystem}} for {{Solving Differential Equations}} in {{Julia}}.
\newblock {\em JORS}, 5(1):15, May 2017.

\bibitem{RamsayHookerCampbellEtAl2007JRStatSocSerBStatMethodol}
J.~O. Ramsay, G.~Hooker, D.~Campbell, and J.~Cao.
\newblock Parameter estimation for differential equations: A generalized smoothing approach.
\newblock {\em J. R. Stat. Soc. Ser. B Stat. Methodol.}, 69(5):741--796, Nov. 2007.

\bibitem{RevelsLubinPapamarkou2016arXiv160707892}
J.~Revels, M.~Lubin, and T.~Papamarkou.
\newblock Forward-{{Mode Automatic Differentiation}} in {{Julia}}.
\newblock {\em arXiv:1607.07892}, July 2016.

\bibitem{RudyBruntonProctorEtAl2017SciAdv}
S.~H. Rudy, S.~L. Brunton, J.~L. Proctor, and J.~N. Kutz.
\newblock Data-driven discovery of partial differential equations.
\newblock {\em Sci. Adv.}, 3(4):e1602614, Apr. 2017.

\bibitem{RussoMessengerBortzEtAl2024IFAC-PapersOnLine}
B.~P. Russo, D.~A. Messenger, D.~M. Bortz, and J.~A. Rosenfeld.
\newblock Weighted {{Composition Operators}} for {{Learning Nonlinear Dynamics}}.
\newblock {\em IFAC-PapersOnLine}, 58(17):97--102, 2024.

\bibitem{Schaeffer2017ProcRSocMathPhysEngSci}
H.~Schaeffer.
\newblock Learning partial differential equations via data discovery and sparse optimization.
\newblock {\em Proc. R. Soc. Math. Phys. Eng. Sci.}, 473(2197):20160446, Jan. 2017.

\bibitem{Shinbrot1954NACATN3288}
M.~Shinbrot.
\newblock On the analysis of linear and nonlinear dynamical systems for transient-response data.
\newblock Technical Report NACA TN 3288, Ames Aeronautical Laboratory, Moffett Field, CA, Dec. 1954.

\bibitem{ShonkwilerHerod2009}
R.~W. Shonkwiler and J.~Herod.
\newblock {\em Mathematical {{Biology}}: {{An Introduction}} with {{Maple}} and {{Matlab}}}.
\newblock Undergraduate {{Texts}} in {{Mathematics}}. Springer New York, New York, NY, 2009.

\bibitem{Sparrow1982}
C.~Sparrow.
\newblock {\em The {{Lorenz Equations}}: {{Bifurcations}}, {{Chaos}}, and {{Strange Attractors}}}, volume~41 of {\em Applied {{Mathematical Sciences}}}.
\newblock Springer, New York, NY, 1982.

\bibitem{Takaya1968IEEETransAutomControl}
K.~Takaya.
\newblock The use of {{Hermite}} functions for system identification.
\newblock {\em IEEE Trans. Autom. Control}, 13(4):446--447, Aug. 1968.

\bibitem{ToniWelchStrelkowaEtAl2009JRSocInterface}
T.~Toni, D.~Welch, N.~Strelkowa, A.~Ipsen, and M.~P. Stumpf.
\newblock Approximate {{Bayesian}} computation scheme for parameter inference and model selection in dynamical systems.
\newblock {\em J. R. Soc. Interface.}, 6(31):187--202, Feb. 2009.

\bibitem{TranHeMessengerEtAl2024ComputMethodsApplMechEng}
A.~Tran, X.~He, D.~A. Messenger, Y.~Choi, and D.~M. Bortz.
\newblock Weak-{{Form Latent Space Dynamics Identification}}.
\newblock {\em Comput. Methods Appl. Mech. Eng.}, 427:116998, July 2024.

\bibitem{WangLiuGibaru2023NonlinearDyna}
L.~Wang, D.-Y. Liu, and O.~Gibaru.
\newblock A new modulating functions-based non-asymptotic state estimation method for fractional-order systems with {{MIMO}}.
\newblock {\em Nonlinear Dyn}, 111(6):5533--5546, Mar. 2023.

\bibitem{WangHuanGarikipati2019ComputMethodsApplMechEng}
Z.~Wang, X.~Huan, and K.~Garikipati.
\newblock Variational system identification of the partial differential equations governing the physics of pattern-formation: {{Inference}} under varying fidelity and noise.
\newblock {\em Comput. Methods Appl. Mech. Eng.}, 356:44--74, Nov. 2019.

\bibitem{WongYangKou2024JStatSoft}
S.~W.~K. Wong, S.~Yang, and S.~C. Kou.
\newblock {\textbf{Magi}} : {{A Package}} for {{Inference}} of {{Dynamic Systems}} from {{Noisy}} and {{Sparse Data}} via {{Manifold-Constrained Gaussian Processes}}.
\newblock {\em J. Stat. Soft.}, 109(4), 2024.

\bibitem{YangWongKou2021ProcNatlAcadSciUSA}
S.~Yang, S.~W.~K. Wong, and S.~C. Kou.
\newblock Inference of dynamic systems from noisy and sparse data via manifold-constrained {{Gaussian}} processes.
\newblock {\em Proc Natl Acad Sci USA}, 118(15):e2020397118, Apr. 2021.

\bibitem{ZhangLuLiuEtAl2021NonlinearDyn}
T.~Zhang, Z.-r. Lu, J.-k. Liu, and G.~Liu.
\newblock Parameter identification of nonlinear systems with time-delay from time-domain data.
\newblock {\em Nonlinear Dyn}, 104(4):4045--4061, June 2021.

\end{thebibliography}
\appendix 
\setcounter{section}{0}
\section{Test Functions}\label{appendix:TestFuns}

As the efficacy of the weak form algorithms depend strongly on wisely choosing test function hyperparameters (smoothness, radius, etc.), here we give an exposition of how we create the test function matrix $\dTestFun \in \mathbb{R}^{K \times M+1}$. The goal is to have minimal hyperparameter tuning specific to a particular dataset, but to rather adaptively select test functions by analyzing the data itself. A similar process was described in \cite{BortzMessengerDukic2023BullMathBiol}.

Starting from Eq.~\eqref{eq:ode-weak}, we must choose a finite set of test functions $\{\varphi_k\}_{k=1}^{K}$ to create the weak residual $\wRes\in \mathbb{R}^K$ where each entry is the weak form EE residual for a particular test function $r_k = \langle \dstate, \dot{\testFun}_k\rangle + \langle \rhs(\dstate), \testFun_k\rangle$. In approaches such as the Finite Element Method, this set is chosen based upon the expected properties of the solution as well as the class of equations being solved. Conversely, to perform weak form EE-based parameter estimation, we assume the availability of samples of the (possibly noisy) solution. In pursuit of the estimation goal, however, we note that how to optimally choose the properties, number, and location of the test functions remains an open problem.
Nonetheless, in \cite{MessengerBortz2021MultiscaleModelSimul}, it was reported that smoother test functions result in a more accurate convolution integral (via the Euler–Maclaurin formula), while later research revealed that test functions with wider support frequently performed better, and subsequently an algorithm to choose the support size was developed in \cite{MessengerBortz2021JComputPhys}. These empirical observations regarding smoothness and support size have partially guided the design of the following algorithm.

We begin by considering the following $\mathcal{C}^\infty$ bump function
\begin{equation}
	\label{eq:test-fun}
	\varphi_k(t ; \eta, m_t)=C \exp \left(-\frac{\eta}{\left[1-(\tfrac{t-t_k}{m_t\Delta t} )^2\right]_{+}}\right)
\end{equation}
where $t_k$ is the center of the function, $m_t$ controls the radius of the support, the constant $C$ normalizes the test function such that $\|\varphi_k\|_2=1$, $\eta$ is a shape parameter, and $[\cdot]_{+}\eqdef\max (\cdot, 0)$, so that $\varphi_k(t ; m_t\Delta t)$ is supported only on $[-m_t\Delta t, m_t\Delta t]$. 
Using the algorithm below in \ref{appendix:minRadiusSelect}, we identify the minimal radius $\underline{m}_t$ such that the numerical error in the convolution integral does not dominate. We desire a set of test functions that are compactly supported over the time domain $(0,T)$, and have the following features: 1) the $t_k$ values coincide with the sampled timepoints and 2) allowable radii are at least as large as the minimal one $m_t \geq \underline{m}_t$. Motivated by empirical results, the shape parameter is arbitrarily fixed at $\eta=9$. This gives us a set of $K_\textrm{full}$ test functions $\{\varphi_k\}_{k=1}^{K_\text{full}}$ and we evaluate them on the grid to obtain the matrices:
\[\dTestFun_\text{full} \eqdef \begin{bmatrix}
	\testFun_1(0) & \testFun_1(\Delta t) & \cdots & \testFun_{1}(T)\\
	\vdots & \vdots & \cdots & \vdots\\
	\testFun_{K_\text{full}}(T) & \testFun_{K_\text{full}}(\Delta t) & \cdots & \testFun_{K_\text{full}}(T)
\end{bmatrix}\in \mathbb{R}^{K \times M+1},\]
\[\dot{\dTestFun}_\text{full} \eqdef \begin{bmatrix}
	\dot{\testFun}_1(0) & \dot{\testFun}_1(\Delta t) & \cdots & \dot{\testFun}_{1}(T)\\
	\vdots & \vdots & \cdots & \vdots\\
	\dot{\testFun}_{K_\text{full}}(T) & \dot{\testFun}_{K_\text{full}}(\Delta t) & \cdots & \dot{\testFun}_{K_\text{full}}(T) 
\end{bmatrix}\in \mathbb{R}^{K \times M+1}.\]

\subsection{Minimum Radius Selection.} \label{appendix:minRadiusSelect}


The analysis presented here follows from the equation $\dot{u}=f(u)$ where $u$ is a one dimensional state variable. The results generalize to a $D$ dimensional system by applying the radius selection in each dimension independently. Using integration by parts for a test function $\varphi$ gives Equation \eqref{eq:ode-weak-trick} $-\langle \dot{\varphi}, u\rangle = \langle \varphi, f(u) \rangle$. Integrating the residual of Equation \eqref{eq:ode-weak-trick} we obtain
\begin{equation} \label{eq:prodRuleStart} 
	0 =  \int_0^T f(u(t))\varphi(t) + \varphi'(t)u(t) dt = \int_0^T \frac{d}{dt}\bigl(u(t)\varphi(t)\bigr) dt. 
\end{equation}
By expanding $\frac{d}{dt}\bigl(u(t) \varphi(t)\bigr)$ into its Fourier Series, where the coefficients are given by $\mathcal{F}_n[\cdot] := \frac{1}{T}\int_0^T (\cdot) \exp\left[\frac{-2\pi i n }{T}t\right] dt$, we can make the following simplification:
\begin{align*} 
	\frac{d}{dt}\bigl(u(t)\varphi(t)\bigr) &= \sum_{n\in\mathbb{Z}} \mathcal{F}_n \left[\frac{d}{dt}\bigl(u(t)\varphi(t)\bigr)\right] \exp \left[\frac{ 2\pi i n}{T}t\right]\\
	&= \sum_{n\in\mathbb{Z}} \left(\frac{1}{T} \int_0^T \frac{d}{dt}\bigl(u(t)\varphi(t)\bigr)  \exp \left[\frac{ -2\pi i n}{T}t\right] dt\right)  \exp \left[\frac{ 2\pi i n}{T}t\right]\\
	&\stackrel{\text{IBP}}{=} \frac{ -2\pi i }{T} \sum_{n\in\mathbb{Z}} n\underbrace{\left( \frac{1}{T} \int_0^T u(t)\varphi(t)  \exp \left[\frac{ -2\pi i n}{T}t\right] dt \right)}_{=\mathcal{F}_n\bigl[\varphi u\bigr]}  \exp \left[\frac{ 2\pi i n}{T}t\right].
\end{align*} 
The last simplification relies that $\varphi$ has compact support on the interior of the time domain $\operatorname{supp}(\varphi) \subset (0,T)$. Plugging this fourier series into Equation \eqref{eq:prodRuleStart} we see that the following sum is equal to zero:
\[0=\int_0^T \frac{d}{dt}\bigl(u(t)\varphi(t)\bigr) \, dt = \sum_{n\in\mathbb{Z}} \mathcal{F}_n\bigl[\varphi u\bigr]\underbrace{\int_0^T \exp \left[\frac{ 2\pi i n}{T}t\right] \, dt}_{\eqdef I_n}.\] 
\begin{remark}[All Elements of the True Sum are Zero]
	Notice that $\forall n \neq 0$, $I_n= 0$ because $\exp\left[\tfrac{2\pi i}{T}t\right]dt$ is a periodic non-constant function on $[0,T]$. Also, $\mathcal{F}_0[\varphi u]= 0$ because it corresponds to the original integral $\mathcal{F}_0[\varphi u] = \mathcal{F}_0\bigl[\frac{d}{dt}(\varphi u)\bigr] = \int_0^T \frac{d}{dt}\bigl(u(t)\varphi(t)\bigr) \, dt$.
\end{remark}
Numerical error is introduced by substituting the evaluation of the integral with a quadrature rule. In this work, we use the trapezoidal rule on a uniform time grid\footnote{This corresponds to integration error defined in Lemma \ref{lemma:int-error}.}.
\begin{equation} \label{eq:eint} 
	e_\text{int} \eqdef \frac{2\pi i}{M} \sum_{n\in\mathbb{Z}} n \mathcal{F}_n\bigl[\varphi u\bigr]\underbrace{\sum_{m=0}^M  \exp \left[\frac{ 2\pi n i }{M}m\right] }_{\eqdef \hat{I}_n}.
\end{equation}
\begin{remark}[Integration Error comes from Specific $\hat{I}_n$]
	 Notice that when $n = \ell M$ for some integer $\ell$, then we have that $\hat{I}_{n} = M$ because the quadrature nodes align with the roots of unity. This motivates inspecting these modes $\mathcal{F}_{\ell M}\bigl[\varphi u\bigr]$ in particular to gauge how the numerical integration error is behaving.
\end{remark}
Because the Fourier modes decay as $n$ gets larger, the error due to numerical integration is dominated by the largest of these Fourier modes $n=M$:
\[e_\text{int} \approx \operatorname{imag}\left\{ 2\pi n \mathcal{F}_M\bigl[\varphi u\bigr]\right\}.\]
Unfortunately, this approximation to the numerical integration error is not explicitly available because $u$ is not known analytically the Fourier coefficients must be approximated by the Discrete Fourier Transform (DFT), $\hat{\mathcal{F}}_{n - \left\lfloor \tfrac{M}{2} \right\rfloor}[\dTrueState] = \frac{T}{M} \sum_{m=0}^{M} \dTrueState_m \exp \left[\frac{ 2\pi i n}{M} m\right]$. Furthermore, the largest mode we can approximate via the DFT is $\hat{\mathcal{F}}_{\left\lfloor\frac{M}{2}\right\rfloor}$. Thus we can find an error estimate of applying the same quadrature rule to a subsampled grid. This gives an upper bound on the numerical integration error: 
\[e_\text{int}(M) \leq e_\text{int}(\tilde{M})\]
where $\tilde{M} \eqdef \bigl\lfloor \frac{M}{s} \bigr\rfloor$ for some scaling factor $s\geq2$. $e_\text{int}(\tilde{M})$ is the integration error when the data is subsampled.

The last complication is that we do not have access to $\dTrueState$, so we must approximate by substituting $\dstate$: 
\[e_\text{int}(\tilde{M}) \leq \hat{e}_\text{int}(\tilde{M}) \approx \frac{2 \pi}{\sqrt{T}} \hat{F}_{\tilde{M}}(\dstate).\]

\begin{figure}[H] 
	\centering
	\includegraphics[width=0.9\textwidth]{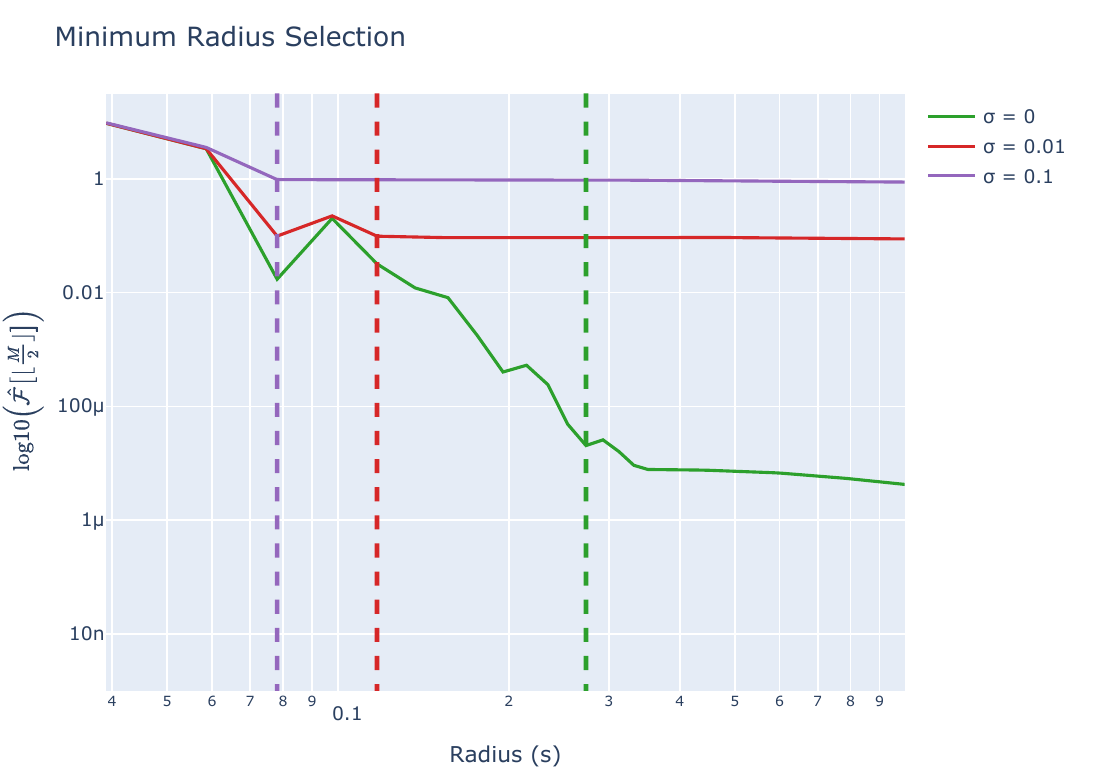}
	\caption{The results for the logistic growth equation $\dot{u} = u + u^2$, on the time domain $(0,10)$, initial condition of $u(0)=0.01$ and $M=512$, looking at $\hat{\mathcal{F}}_{256}$ for a variety of radii  $\underline{m_t} \in [0.04, 1]$. A vertical line indicates where we have detected a corner in the integration error surrogate.}
	\label{fig:minRad}
\end{figure}

We use this bound to search over the space of possible radii. Empirically, we have found that there is threshold for the radius where the effects due to noise have yet to dominate. We found that as the noise level increases the effects from noise dominate for smaller radii, thus for higher noise we expect to select smaller minimum radius. This can be seen in Fig.~\ref{fig:minRad}.
The goal of this procedure is to find the corner shown in Fig \ref{fig:minRad}. Choosing radii for the test functions larger than the minimum radius $m_t \geq \underline{m_t}$ allows for the test functions to preserve as much information about the state as possible while damping the effects due to noise. This leads to better downstream performance in the weak form algorithms.
\subsection{Improving the Conditioning of the Test Function Matrix}
\newcommand{\leftSingVals}{\mathbf{W}}
\newcommand{\rightSingVals}{\mathbf{V}}
The test function matrix $\dTestFun_\text{full}$ is often overdetermined, so in order to improve the conditioning of the system we use an SVD reduction technique to obtain a matrix $\dTestFun$ that has a better condition number $\kappa(\dTestFun) \eqdef \sigma_1(\dTestFun)  / \sigma_K(\dTestFun)$, while still retaining as much information as possible. This is done by looking for a corner in the singular values.
Let the SVD of $\dTestFun_\text{full}$ by 
\begin{align*} 
	\dTestFun_\text{full} &= \leftSingVals \operatorname{diag}([\sigma_1, \cdots, \sigma_K, \cdots]) \rightSingVals^\top  
	\shortintertext{ where $\leftSingVals, \rightSingVals$ are unitary and $\sigma_1 \ge \sigma_2 \ge \ldots \ge 0$.
    We then define} \\
	\dTestFun &= \underbrace{\operatorname{diag}\Bigl(\bigl[\tfrac{1}{\sigma_1}, \cdots, \tfrac{1}{\sigma_K}\bigr]\Bigr)\leftSingVals^\top}_{\mathbf{P}} \dTestFun_\text{full} 
\end{align*} 
where the cutoff $K$ is found by looking for a corner in the plot of $\sigma_i$ vs $i$ or by setting a desired information threshold. \begin{remark}[Functional Interpretation]
	Numerically, this technique finds a set orthonormal test functions that span a dominate subspace of the space spanned by the original test functions.
\end{remark}
This improves on the approach in \citep{BortzMessengerDukic2023BullMathBiol} which approximated $\dot{\dTestFun}$ with a spectral method. In contrast, we compute  $\dot{\dTestFun}_\text{full}$ analytically, and then leverage the SVD of $\dTestFun$ to apply the same linear operators, $\dot{\dTestFun}=\mathbf{P}\dot{\dTestFun}_\text{full}$, to obtain a result that only has error from numerical precision,
leading to slightly better numerical integration error.

\section{Proofs}\label{appendix:proofs}

Two Lemmas are proved as they are necessary for the proofs of the larger results. Then the proof of Proposition \ref{prop:wres-dist} is given, followed by Corollary \ref{cor:log-dist}.

\begin{lemma} \label{lemma:int-error}
	Let uncorrupted data $\dTrueStateMat$ and true parameters $\trueParams$ satisfy Equation \eqref{eq:ode-weak} on the time domain $[0,T]$. Assuming that $\rhs$ is continuous in time, the following holds:
	\[\lim_{M\rightarrow \infty} \bigl\|\wResQuad(\trueParams)\bigr\| = 0\]
	where \(\wResQuad(\trueParams) = \wRhs(\trueParams; \dTrueStateMat,\dt) -\wLhs(\dTrueStateMat)\)
\end{lemma}
\begin{proof}
	The uncorrupted data (true state) satisfies the weak form of the system of differential equations as stated in Equation \eqref{eq:ode-weak} for all possible test functions. The $k^\text{th}$ entry of $\wResQuad$ is an approximation of Equation \eqref{eq:ode-weak} for a particular test function $\testFun_k$:
	\[r^{\text{int}}_k(\trueParams) = \underbrace{\langle \testFun_k, \rhs(\trueParams, \ustate, \t)\rangle + \langle \dot{\testFun_k}, \ustate \rangle}_{=0} + \mathbf{e}^{\text{int} }\] 
	The inner products are computed via a quadrature rule. Relying on the assumption that $\ustate$ is in a Sobolev space, for a fixed time domain the integration error, $\mathbf{e}^\text{int}$, converges to zero as the number of points $M$ increases:
	\[\lim_{M\rightarrow\infty} \|\wResQuad(\trueParams)\| = \lim_{M\rightarrow\infty} \|\mathbf{e}^{\text{int}}\| = 0.\]
\end{proof}

\begin{lemma} \label{lemma:noise}
	For uncorrupted  data $\dTrueStateMat$ and true parameters $\trueParams$ that satisfy Equation \eqref{eq:ode-weak}, we define the corrupted data as $\dstatemat = \dTrueStateMat + \dNoise$, where $\operatorname{vec}[\dNoise] \sim \mathcal{N}(\mathbf{0},  \mathbf{\Sigma} \otimes \id_{M+1})$. Assuming that $\rhs$ is twice continuously differentiable in $\ustate$, the following holds:
	\[\wCov(\trueParams)^{-\tfrac{1}{2}} \wResLin(\trueParams) \sim \mathcal{N}({\mathbf{0}},\id_{KD}) \]
	and
	\[\mathbb{E}\Bigl[\bigl\|\wResNoise(\trueParams) - \wResLin(\trueParams)\bigr\|\Bigr] = \mathcal{O}\Bigl(\mathbb{E}\bigl[\|\dNoise\|^2\bigr]
	\Bigr) \] where 
	\begin{align*}
		\wResLin(\trueParams) &\eqdef \nabla_{\ustate} \wRhs(\trueParams;\dstatemat,\dt)\operatorname{vec}[\dNoise] - \wLhs(\dNoise), \\
		\wResNoise(\trueParams) &\eqdef \wRhs(\trueParams; \dstatemat, \dt) - \wRhs(\trueParams; \dTrueStateMat,\dt) -\wLhs(\dNoise),\\
		\wCov(\trueParams) &\eqdef \Bigl(\nabla_{\ustate} \wRhs(\trueParams;\dstatemat,\dt) + (\id_D \otimes \dot{\dTestFun}) \Bigr) \Bigl(  \mathbf{\Sigma} \otimes \id_{M+1} \Bigr) \\
		&\quad\;\;\;\Bigl(\bigl(\nabla_{\ustate} \wRhs(\trueParams;\dstatemat,\dt)\bigr)^\top + (\id_D \otimes \dot{\dTestFun}^\top)\Bigr),
	\end{align*}
	and $\nabla_{\ustate} \wRhs(\trueParams;\dstatemat,\dt) \in \R^{KD \times D(M+1)}$ is the Jacobian matrix of $\wRhs(\trueParams;\dstatemat,\dt)$ with respect to the state variable, $\ustate$.
\end{lemma}

\begin{proof}
	First, the distribution of $\wResLin$ follows from the Gaussian distribution of $\dNoise$. Since a linear combination of  jointly Gaussian random variables (with a positive definite covariance matrix) is also Gaussian, we can compute the following mean and covariance conditioned on the observed data:
	\begin{align*}
		\mathbb{E}\bigl[\wResLin(\trueParams) \mid \dstatemat\bigr] &= 0, \\ 
		\mathbb{E}\Bigl[\wResLin(\trueParams) \bigl(\wResLin(\trueParams)\bigr)^\top \mid \dstatemat\Bigr] &= \Bigl(\nabla_{\ustate} \wRhs(\trueParams;\dstatemat,\dt) + (\id_D \otimes \dot{\dTestFun} ) \Bigr) \Bigl(  \mathbf{\Sigma} \otimes \id_{M+1} \Bigr) \\
		&\quad\;\;\Bigl(\bigl(\nabla_{\ustate} \wRhs(\trueParams;\dstatemat,\dt)\bigr)^\top + (\id_D \otimes \dot{\dTestFun}^\top )\Bigr).
	\end{align*}
	To obtain the expected error from $\wResLin$ to $\wResNoise$, we begin by expanding $\wResNoise$ about the data $\dstatemat$. Taylor's Remainder Theorem guarantees that $\exists \mathbf{\hat{U}} \in \bigl\{X \in \mathbb{R}^{(M+1)\times D} \mid \|X - U\|_\infty \leq \|\dstatemat-\dTrueStateMat\|_\infty\bigr\}$ such that
	\begin{align*}
		\wResNoise(\trueParams) &= \wRhs(\trueParams; \dstatemat, \dt) - \wLhs(\dNoise)- \bigl(\wRhs(\trueParams;\dstatemat,\dt) - \nabla_{\ustate} \wRhs(\trueParams;\dstatemat,\dt)\operatorname{vec}[\dNoise] \\
		&\quad + \nicefrac{1}{2}(\mathbf{H}_{\ustate}(\wRhs)(\trueParams; \mathbf{\hat{U}}, \dt) \bar{\times}_3 \operatorname{vec}[\dNoise]) \bar{\times}_2 \operatorname{vec}[\dNoise] \bigr) 
	\end{align*}
	where $\mathbf{H}_{\ustate}(\wRhs): \mathbb{R}^J \times \mathbb{R}^{(M+1) \times D} \times \mathbb{R}^{M+1} \rightarrow \mathbb{R}^{KD\times (M+1)D \times (M+1)D}$ is the Hessian of $\wRhs$ with respect to $\ustate$, and $\bar{\times}_n$ specifies the \emph{n-mode (vector) product} of a tensor with a vector (see section 2.5 of \citep{KoldaBader2009SIAMRev}. The $\wRhs(\trueParams; \dstatemat, \dt)$ terms cancel, leaving
	\[\wResNoise(\trueParams) = \underbrace{\nabla_{\ustate} \wRhs(\trueParams;\dstatemat,\dt)\operatorname{vec}[\dNoise] - \wLhs(\dNoise)}_{=\wResLin(\trueParams)} + \nicefrac{1}{2} (\mathbf{H}_{\ustate}(\wRhs)(\trueParams; \mathbf{\hat{U}}, \dt) \bar{\times}_3 \operatorname{vec}[\dNoise]) \bar{\times}_2 \operatorname{vec}[\dNoise].\]
	Subtracting $\wResLin$, taking the expectation of the norm of both sides, and then applying triangle inequality simplifies to the desired result
	\[\mathbb{E}\Bigl[\bigl\|\wResNoise(\trueParams) - \wResLin(\trueParams)\bigr\|\Bigr] = \mathcal{O}\Bigl(\mathbb{E}\bigl[\|\dNoise\|^2\bigr]\Bigr).\]
\end{proof}

\primeProb*
\begin{proof}
	Regardless of the form of $\rhs$, we have that $\wLhs$ is linear, thus:
	\[\wLhs(\dstatemat) = \wLhs(\dTrueStateMat) + \wLhs(\dNoise)\]
	Next, for  true parameters $\trueParams$, we have:
	\begin{align*}
	\wRes(\trueParams;\dstatemat,\dt) &= \wRhs(\trueParams;\dstatemat,\dt) - \wRhs(\trueParams;\dTrueStateMat,\dt) + \wRhs(\trueParams; \dTrueStateMat,\dt) - \wLhs(\dTrueStateMat) - \wLhs(\dNoise)
	\end{align*}
	Regrouping terms isolates the sources of error in the weak residual:
	\begin{align*}
		\wRes(\trueParams;\dstatemat,\dt) &= \underbrace{\wRhs(\trueParams; \dTrueStateMat,\dt) - \wLhs(\dTrueStateMat)}_{= \wResQuad(\trueParams)} + \underbrace{\wRhs(\trueParams; \dstatemat, \dt) - \wRhs(\trueParams; \dTrueStateMat,\dt) -\wLhs(\dNoise)}_{= \wResNoise(\trueParams)}\\
		\lim_{M\rightarrow \infty} \mathbb{E}\Bigl( \|\wRes(\trueParams) - \wResLin(\trueParams)\| \Bigr) &\leq \lim_{M\rightarrow \infty} \mathbb{E}\Bigl( \|\wResQuad(\trueParams)\| \Bigr) + \mathbb{E}\Bigl( \|\wResNoise(\trueParams) - \wResLin(\trueParams)\|\Bigr)
	\end{align*}
	Combining the results from Lemmas \ref{lemma:int-error} and \ref{lemma:noise}, the result holds. 
\end{proof}

\logCor*
where $\oslash$ is element-wise division and
\[\begin{aligned}
    \logRhs(\params, \dstatemat, \dt) &\eqdef \rhs(\params, \ustate, \t) \oslash \ustate, \;
    \wlogRhs(\params;\dstatemat,\dt) \eqdef \operatorname{vec}\bigl[\dTestFun \logRhs(\params, \dstatemat, \dt)\bigr]\\
    \wlogLhs(\dstatemat) &\eqdef \operatorname{vec} \bigl[- \dot{\dTestFun} \log(\dstatemat)\bigr],\;
    \wlogRes \eqdef \wlogRhs(\params;\dstatemat,\dt) - \wlogLhs(\dstatemat), \\
    \wlogResLin &\eqdef \nabla_{\ustate} \wlogRhs(\trueParams; \dstatemat, \dt)\operatorname{vec}[\log(\dlognoise)] - \wlogLhs(\dlognoise), \\
    \wlogCov(\params;\dstatemat,\dt) &\eqdef \bigl(\nabla_{\log(\ustate)} \wlogRhs(\params;\dstatemat,\dt) + \dot{\dTestFun}\bigr)\bigl( \mathbf{\Sigma} \otimes \id_{M+1}\bigr)\bigl(\nabla_{\log(\ustate)}\wlogRhs(\params;\dstatemat,\dt)+ \dot{\dTestFun}\bigr)^\top
\end{aligned}\]
\begin{proof}
	Start by defining a change of variables:
	\[\{\dlogstate_m\}_{m=0}^M \eqdef \{\log(\dstate_m)\}_{m=0}^M = \{\log(\dTrueState_m) + \log(\lognoise_m)\}_{m=0}^M.\]
	We build the corresponding data matrix \(\dlogstatemat \eqdef \begin{bmatrix}
		\dlogstate_0 & \cdots & \dlogstate_M
	\end{bmatrix}\).
	Applying chain rule we see that $\logstate$ satisfies the following ODE:
	\[\frac{d\logstate}{dt} = \rhs(\params, \exp(\logstate), \t) \oslash \exp(\logstate) \]
	Thus, we have the corresponding weak form of the ODE for transformed state variable:
	\[ -\langle \dot{\testFun}, \logstate \rangle = \langle \testFun , \rhs(\params, \exp(\logstate), \t) \oslash \exp(\logstate)\rangle. \]
	Notice that for $\dlogstatemat$, the noise is now additive and Gaussian, thus we satisfy the assumptions of Proposition \ref{prop:wres-dist}, and can obtain the desired result.
\end{proof}

\begin{remark}[Derivative Computations for the Log-Normal Case]
	In practice, the computations for the RHS, LHS and corresponding derivatives stay identically the same, modulo the replacing of $\wLhs$ with $\wlogLhs$ and $\rhs$ with $\logRhs$. The computations for $\partial_{p_j} \nabla_{\ustate}  \wRhs(\params;\dstatemat,\dt)$ and $\partial_{p_i,p_j}\nabla_{\ustate}  \wRhs(\params;\dstatemat,\dt)$ now become $\partial_{p_j} \nabla_{\logstate}\wRhs(\params;\exp(\dlogstatemat))$ and  $\partial_{p_i,p_j}\nabla_{\logstate} \wRhs(\params;\exp(\dlogstatemat))$, respectively.
\end{remark}

\section{Derivative Information of the Weak Likelihood}\label{appendix:DerInfo}
First and second order derivative information of $\loglikelihood$ can be derived analytically, allowing for efficient use of second-order optimization routines that are robust to non-convex problems. All the derivative computations assume $\rhs$ is twice continuously differentiable in $\params$. For ease of notation and because the data is fixed, we drop explicit dependence on $\dstatemat$ and $\dt$: $\wRhs(\params) \eqdef \wRhs(\params;\dstatemat,\dt)$, and $\wLhs \eqdef \wLhs(\dstatemat)$.

\subsection{Gradient Information}

Regardless of the linearity of $\rhs$ with respect to $\params$, the $j^\text{th}$ component of the gradient of the weak log-likelihood is given as 
\begin{equation} \label{eq:wnll-grad}
	\begin{aligned}
		\partial_{p_j} \loglikelihood(\params) &= \frac{1}{2} \Bigl( \operatorname{Tr}\bigl(\wCov(\params)^{-1}\partial_{p_j}\wCov(\params)\bigr) \\
		&+ 2 \bigl(\partial_{p_j} \wRhs(\params)\bigr)^\top \wCov(\params)^{-1} \bigl(\wRhs(\params)- \wLhs\bigr) \\
		&+ \bigl(\wRhs(\params)-\wLhs\bigr)^\top\bigl(\partial_{p_j} \wCov(\params)^{-1}\bigr)\bigl(\wRhs(\params)-\wLhs\bigr)\Bigr)
	\end{aligned}
\end{equation}
where
\begin{align*}
	\partial_{p_j} \wCov(\params)^{-1} &= -\wCov(\params)^{-1} \left(\partial_{p_j}\wCov(\params) \right) \wCov(\params)^{-1},\\
	\partial_{p_j}\wCov(\params) &= \partial_{p_j} \nabla_{\ustate} \wRhs(\params) \bigl(  \mathbf{\Sigma} \otimes \id_{M+1} \bigr)\bigl(\nabla_{\ustate} \wRhs(\params) + \dot{\dTestFun} \otimes \id_D \bigr)^\top \\
	&+ \bigl(\nabla_{\ustate} \wRhs(\params) + \dot{\dTestFun} \otimes \id_D \bigr)\bigl(  \mathbf{\Sigma} \otimes \id_{M+1} \bigr)\partial_{p_j} \nabla_{\ustate} \wRhs(\params) ^\top .
\end{align*}
\subsection{Hessian Information}
Taking derivatives again, the elements of Hessian are
\begin{equation}
	\begin{aligned} \label{eq:wnll-hess}
		\partial_{p_i p_j} \loglikelihood(\params) &= \frac{1}{2}\Bigl( \operatorname{Tr}\bigl(\partial_{p_i}\wCov(\params)^{-1} \partial_{p_j}\wCov(\params) + \wCov(\params)^{-1} \partial_{p_i,p_j}\wCov(\params)\bigr) \\
		&+ 2\bigl(\partial_{p_j} \wRhs(\params)\bigr)^\top \bigl(\partial_{p_i} \wCov(\params)^{-1}\bigr) \bigl(\wRhs(\params) - \wLhs\bigr) \\
		&+ 2\bigl(\partial_{p_i} \wRhs(\params)\bigr)^\top \bigl(\partial_{p_j} \wCov(\params)^{-1}\bigr) \bigl(\wRhs(\params) - \wLhs\bigr) \\
		&+ 2\bigl(\partial_{p_j} \wRhs(\params)\bigr)^\top \wCov(\params)^{-1} \partial_{p_i} \wRhs(\params) \\
		&+ 2\bigl(\partial_{p_i,p_j} \wRhs(\params)\bigr)^\top \wCov(\params)^{-1} \bigl(\wRhs(\params) - \wLhs\bigr)\\
		&+ \bigl(\wRhs(\params) - \wLhs\bigr)^\top\bigl(\partial_{p_i,p_j} \wCov(\params)^{-1}\bigr)\bigl(\wRhs(\params) - \wLhs\bigr)\Bigr)
	\end{aligned}
\end{equation}
where
\begin{align*}
	\partial_{p_i p_j} \wCov(\params)^{-1} &= \wCov(\params)^{-1} \partial_{p_i} \wCov(\params) \wCov(\params)^{-1} \partial_{p_j} \wCov(\params) \wCov(\params)^{-1} \\
	&- \wCov(\params)^{-1} \partial_{ij} \wCov(\params) \wCov(\params)^{-1} \\
	&+ \wCov(\params)^{-1} \partial_{p_j} \wCov(\params) \wCov(\params)^{-1}  \partial_{p_i} \wCov(\params) \wCov(\params)^{-1}\\
	\partial_{p_i p_j} \wCov(\params)&= \bigl(\partial_{p_i,p_j} \nabla_{\ustate} \wRhs(\params)\bigr) \bigl( \mathbf{\Sigma} \otimes \id_{M+1}\bigr) \bigl(\nabla_{\ustate} \wRhs(\params) + \dot{\dTestFun} \otimes \id_D\bigr)^\top \\
	&+ \bigl(\partial_{p_j} \nabla_{\ustate} \wRhs(\params)\bigr) \bigl( \mathbf{\Sigma} \otimes \id_{M+1}\bigr) \bigl(\partial_{p_i} \wRhs(\params)\bigr)^\top.
\end{align*}

\section{Highlighting the Effects of Linearity in Parameters} \label{appendix:linvsnonlin}
When $\rhs$ is linear in $\params$ then then we can write the residual as follows: 
\begin{equation}
	\label{eq:resLin}
	\wRes(\params;\dstatemat,\dt) = \wRhsLin(\dstatemat, \dt) \params - \wLhs(\dstatemat)
\end{equation}
where $\wRhsLin(\dstatemat, \dt) \in \R^{KD\times J}$ is a matrix-valued function that is constant with respect to $\params$. This leads to simplifications in the derivative information
which can improve computational efficiency. 

In the linear case, the Jacobian of the weak form right-hand side with respect to the parameters is the matrix $\wRhsLin(\dstatemat, \dt)$. Formally, $\nabla_{\params} \wRhs(\params; \dstatemat, \dt) = \wRhsLin(\dstatemat,\dt)$. This also simplifies implies $\nabla_{\params} \nabla_{u} \wRhs(\params; \dstatemat, \dt) = \nabla_{u} \wRhsLin(\dstatemat, \dt)$. We again drop explicit dependence on $\dstatemat$ and $\dt$ for simplicity of notation. Feeding this into our existing expression for the gradient of the weak form negative logarithm, we have
\begin{align*}
	\nabla_{\params} \ell(\params) &= 2 \wRhsLin^\top \wCov(\params)^{-1} (\wRhsLin\params- \wLhs) \\
	&+ (\wRhsLin\params-\wLhs)^\top(\partial_{p_j} \wCov(\params)^{-1})(\wRhsLin\params-\wLhs).
\end{align*}
Also, notice that $\wCov(\params)$ now becomes quadratic in $\params$:
\[\wCov(\params) = \bigl(\nabla_{u} \wRhsLin [\params]+\dot{\dTestFun} \otimes \id_D\bigr) \bigl(\Sigma \otimes \id_{M+1} \bigr)\bigl(\nabla_{u} \wRhsLin [\params] + \dot{\dTestFun} \otimes \id_D\bigr)^\top \]
In fact $\nabla_{u} \wRhsLin$ is the Jacobian of a matrix-valued function, and thus it is a three-dimensional tensor in $\R^{KD \times MD \times J}$. We treat it  as a linear operator acting on $\params$.  In practice this is a page-wise mat-vec across the third dimension of $\nabla_{u} \wRhsLin$. 

This causes the following simplification in the derivative of $\wCov(\params)$:
\[\partial_{p_j} \wCov(\params) = 2\nabla_{u} \wRhsLin [\mathbf{e}_j] \bigl(\Sigma \otimes \id_{M+1} \bigr) \bigl(\nabla_{u}\wRhsLin[\params]+\dot{\dTestFun} \otimes \id_D\bigr)^\top \]
where $\mathbf{e}_j \in \R^J$ is the $j^\text{th}$ canonical basis vector. This is equivalent to indexing into the $j^\text{th}$ page of $\nabla_{u}\wRhsLin$. 

The second order derivative computations also simplify beyond the evaluation of $\wCov(\params)$. Critically, observe that several terms in the Hessian are now guaranteed to be $\mathbf{0}$. In particular, we have $\forall \params$:
\[\partial_{p_i,p_j} \wRhsLin \params = \mathbf{0} \in \R^J \text{ and }  
\partial_{p_i,p_j} \nabla_{u} \wRhsLin[\params] = \mathbf{0} \in \R^{KD\times MD}. \]
Furthermore because $\nabla_{u} \wRhsLin$ is constant with respect to $\params$, then $\partial_{p_i p_j}\wCov(\params)^{-1}$ is constant with respect to $\params$, so this can be computed once and then reused. 

\section{Other Weak Form Methods}\label{appendix:otherWeak}
\subsection{Weak Form Least Squares}
The simplest weak form method is to simply minimize the norm of the weak form residual. We call this algorithm WLS:
\[\estim^{(0)} = \underset{\params \in \R^J}{\operatorname{argmin}} \tfrac{1}{2}\|\wRhs(\params) - \wLhs\|^2_2\]
When the ODE is LiP, the solution to this optimization problem is done explicitly through the linear algebra. In the NiP case, this can no longer be done, so an initial guess is necessary. In this work, we pass all algorithms the same initial guess that is randomly sampled from parameter range specified in Table \ref{tab:odes}.

\subsection{WENDy via Iterative Re-weighted Least Squares}\label{appendix:WENDy-IRLS}

In the previous work \citep{BortzMessengerDukic2023BullMathBiol}, the covariance information was incorporated by solving the generalized least squares problem 
\[ \estim = \underset{\params \in \R^J}{\operatorname{argmin}}\; \wRes(\params)^\top\wCov(\trueParams)^{-1}\wRes(\params).\]
Because $\wCov(\trueParams)$ is not known, it has to be approximated with the current value of $\params$. This gives rise the iterative re-weighted least squares algorithm, which iterates as follows 
\begin{equation}
	\label{eq:WENDy-IRLSIter}
	\estim^{(i+1)} = \underset{\params \in \R^J}{\operatorname{argmin}} \; \wRes(\params)^\top\wCov(\estim^{(i)})^{-1}\wRes(\params) 
\end{equation}
until the iterates are sufficiently close together (and again, we drop explicit dependence on $\dstatemat$ and $\dt$). This iteration involves computing an estimate for the covariance by using the previous iteration's parameters $\wCov(\estim^{(i)})$. Then we solve a weighted least squares problem. In the nonlinear case, we have extended this algorithm by solving a nonlinear weighted least squares problem at each iteration where the Jacobian $\nabla_\params \wRes$ is computed using analytic derivative information. 

\section{Computational Cost}\label{appendix:compCost}

Recall that $K$ is the number of test functions, $M+1$ is the number of time points, $D$ is the dimension of the state variable $\ustate$, and $J$ is the dimension of the parameters $\params$. Computing the weak form negative log-likelihood itself has a complexity of $\mathcal{O}(K^2D^3(MJ + K))$. Similarly, the gradient computation has a complexity of $\mathcal{O}\bigl(K^2D^2\bigl(D(MJ + K)+J\bigr)\bigr)$ and the Hessian computation has a complexity of $\mathcal{O}\bigl(K^2D^2\bigl(D(MJ + K)+J^2\bigr)\bigr)$. The most expensive step in our method is computing the derivatives of the weak form log-likelihood for the second order optimization methods. Our implementation meticulously caches as much as possible, pre-allocates memory, applies an advantageous order of operations, and uses efficient data structures. 

\begin{figure}[H]
    \centering
\begin{subfigure}{0.8\textwidth}
    \includegraphics[width=0.95\linewidth]{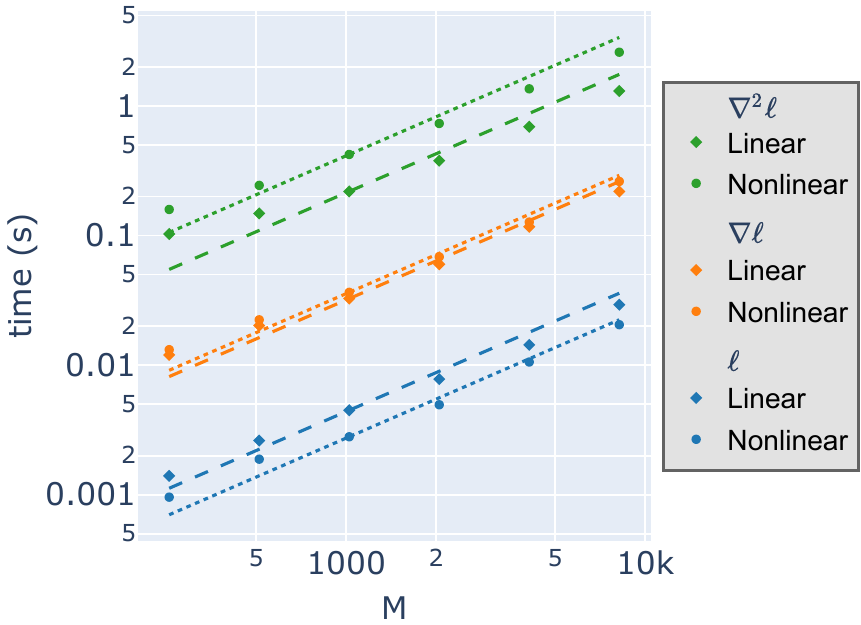}
    \caption{}
    \label{fig:compCost:subM}
\end{subfigure} \\
\begin{subfigure}{0.8\textwidth}
    \includegraphics[width=0.95\linewidth]{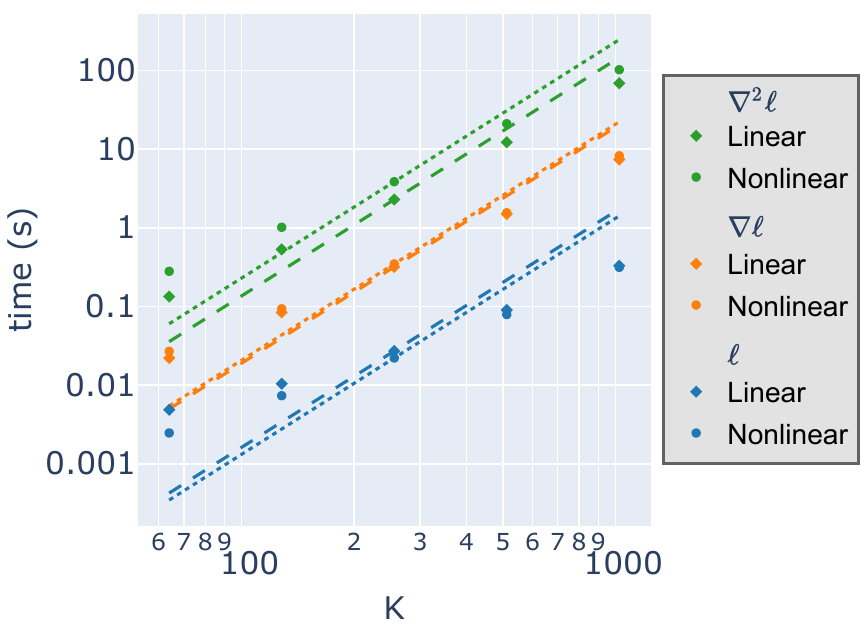}
    \caption{}
    \label{fig:compCost:subK}
\end{subfigure}
\caption{On the top, $K, D, J$ are all fixed, and $M$ is varied. On the bottom, $M, D, J$ are all fixed, and $K$ is varied.  The dashed and dotted lines correspond to a linear fit with a fixed slope ($\mathcal{O}(M), \mathcal{O}(K^3)$ respectively). Both plots are show on a log-log scale.}
\label{fig:compCost}
\end{figure}

In the linear case, many of the operators either become constant matrices or tensors that can be thought of as linear operators. Also, some operators are zero due to higher order derivatives being applied to the linear function. This reduces the constants involved in the cost, but incurs the same asymptotic complexity as the nonlinear case. One can see that the dominant cause for the cost is due to the number of test functions in Fig.~\ref{fig:compCost}. Notice that the recorded run times shown in Fig.~\ref{fig:compCost:subM} approximately match the predicted cost of the line of slope 1. 

The example above has ten parameters ($J=10$). Because the cost is dominated by the mat-mat computations for matrices of size $KD$, the gradient computation takes approximately an order of magnitude more time, and the Hessian takes an order of magnitude more than the gradient. 

Fig.~\ref{fig:compCost} shows the computational cost for different sized problems. Problems that are linear in parameters show a decrease in the computational cost but only by a scaling factor. The experiment verifies the cost of the algorithm scales linearly with respect to the number of data points on the grid and cubically with respect to the number of test functions. Unlike OE methods, the weak form methods can control the computational cost by using fewer test functions. In practice, we used 200-500 test functions which results in reasonable computational cost for the likelihood and its derivatives.

\section{Supplemental Material} \label{appendix:sup}

\subsection{Additional Experimental Details}
\begin{minipage}{\linewidth}
	\centering
\begin{tabular}{cc}
	\toprule 
	 \textbf{Name} &  \textbf{Initial Parameterization} \\
	\midrule 
	\\[-5pt]
	Lorenz & \begin{minipage}{4cm} \(\begin{aligned}
		p_1&\in [0,20]\\
		p_2&\in [0,35]\\
		p_3&\in [0, 5]
	\end{aligned}\) \end{minipage} \\\\[-5pt]
	\hline 
	\\[-5pt]
	Hindmarsh-Rose & \begin{minipage}{4cm} \(\begin{aligned}
		p_1&\in[0,20],\;p_2\in[0,20]\\
		p_3&\in[0,60],\;p_4\in[0,20]\\
		p_5&\in[0,20],\;p_6\in[0,100]\\
		p_7&\in[0,20],\;p_8\in[0,1]\\
		p_9&\in[0,1],\;p_{10}\in[0,1]
	\end{aligned}\) \end{minipage} \\\\[-5pt]
	\hline 
	\\[-5pt]
	Goodwin  & \begin{minipage}{4cm} \(\begin{aligned}
		p_1 &\in [1,5], \; p_2\in[0,0.2] \\
		p_3 &\in [0,2], \; p_4\in[5, 15] \\
		p_5 &\in [0,0.2], \; p_6\in[0,0.2] \\
		p_7 &\in [0,0.2], \; p_8\in[0,0.2] \\
	\end{aligned}\) \end{minipage}\\\\[-10pt]
	\hline 
	\\[-5pt]
	SIR-TDI & \begin{minipage}{4cm}\( \begin{aligned}
		p_1 &\in [10^{-4},1], \; p_2 \in [10^{-4},2] \\
		p_3 &\in [10^{-4},1], \; p_4 \in [10^{-4},1] \\
		p_5 &\in [10^{-4},1] \\
	\end{aligned}\) \end{minipage}\\\\[-5pt]
	\bottomrule
\end{tabular}
\captionof{table}{Table of the ranges that initial parameters are sampled from uniformly.}
\label{tab:initParams}
\end{minipage}

\subsection{Additional Accuracy Plots}
Here we show plots similar to that in the results section, but distinguish between different subsampling levels so that readers can see how each algorithm behaves as the data becomes more sparse.
\begin{figure}[H]
	\centering
	\includegraphics[width=1\textwidth]{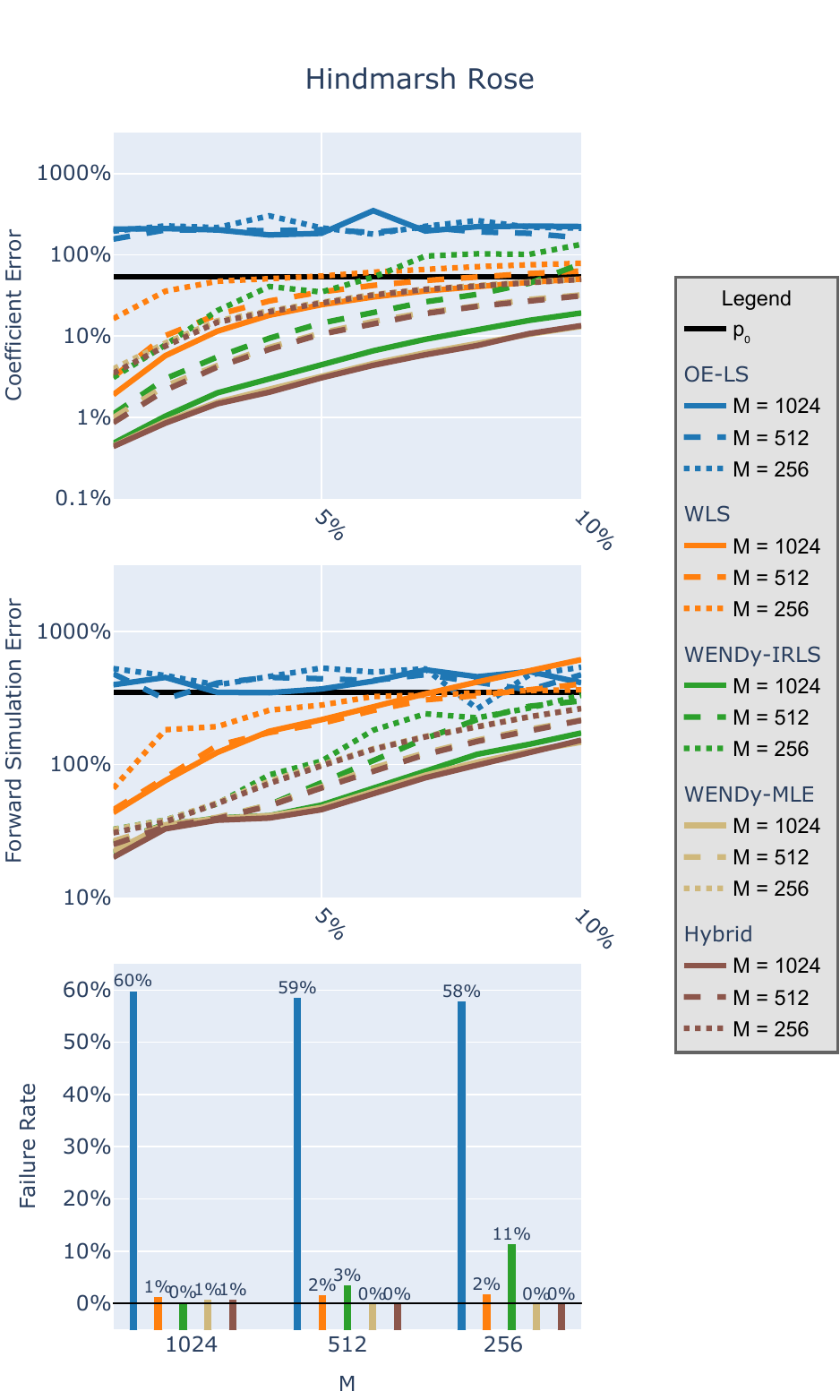}
	\caption{All algorithms are run at noise ratios from 1\% to 20\% and $M = \{256, 512, 1024\}$. We see accuracy metrics for for OE-LS in blue, WLS in orange, WENDy-IRLS in green, WENDy-MLE in gold, the hybrid method in brown, and initial parameterization in black.} 
	\label{fig:hindmarshFleshedOut}
\end{figure}

\begin{figure}[H]
	\centering
	\includegraphics[width=1\textwidth]{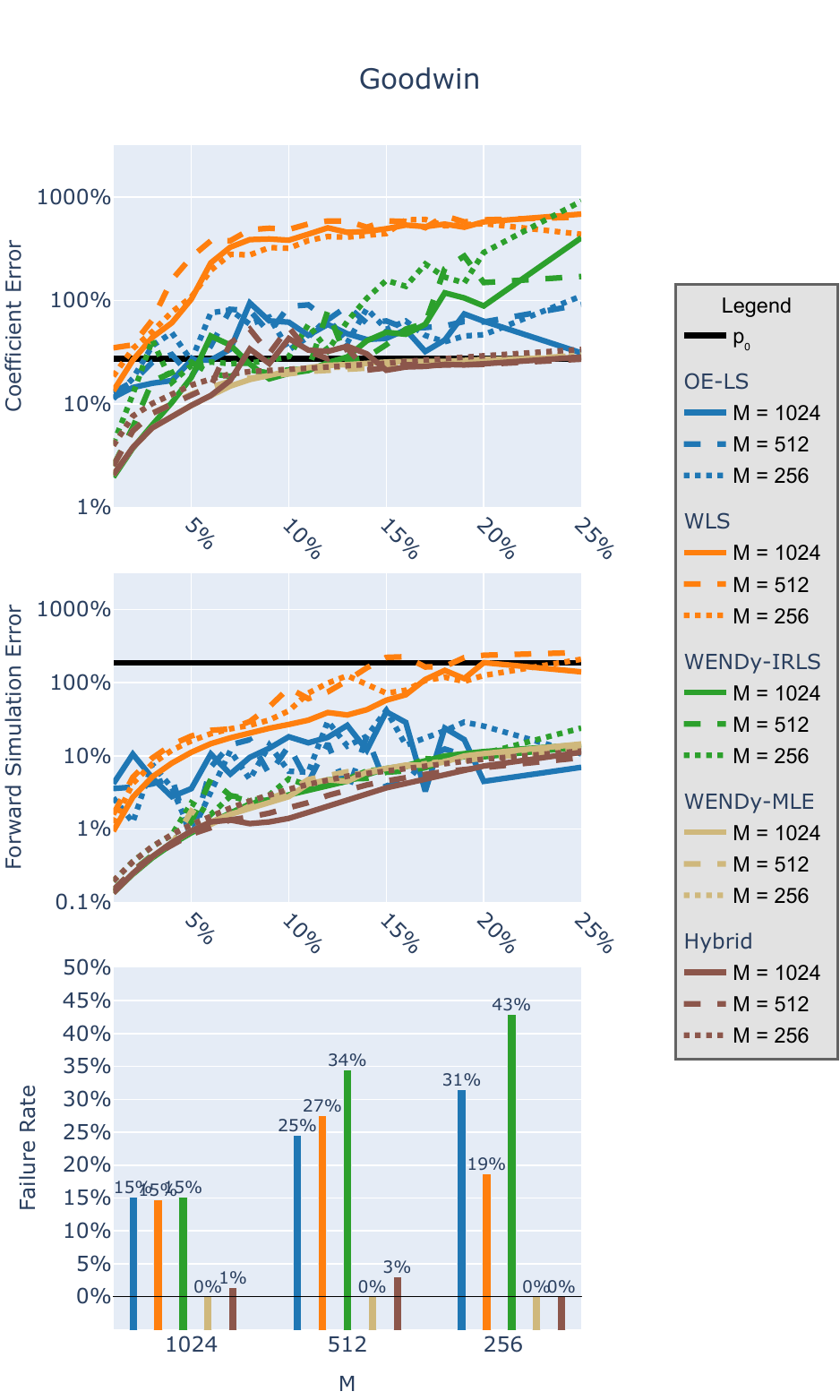}
	\caption{All algorithms are run at noise ratios from 1\% to 25\% and $M = \{256, 512, 1024\}$. We see accuracy metrics for for OE-LS in blue, WLS in orange, WENDy-IRLS in green, WENDy-MLE in gold, the hybrid method in brown, and initial parameterization in black.} 
	\label{fig:goodwinFleshedOut}
\end{figure}

\begin{figure}[H]
	\centering
	\includegraphics[width=1\textwidth]{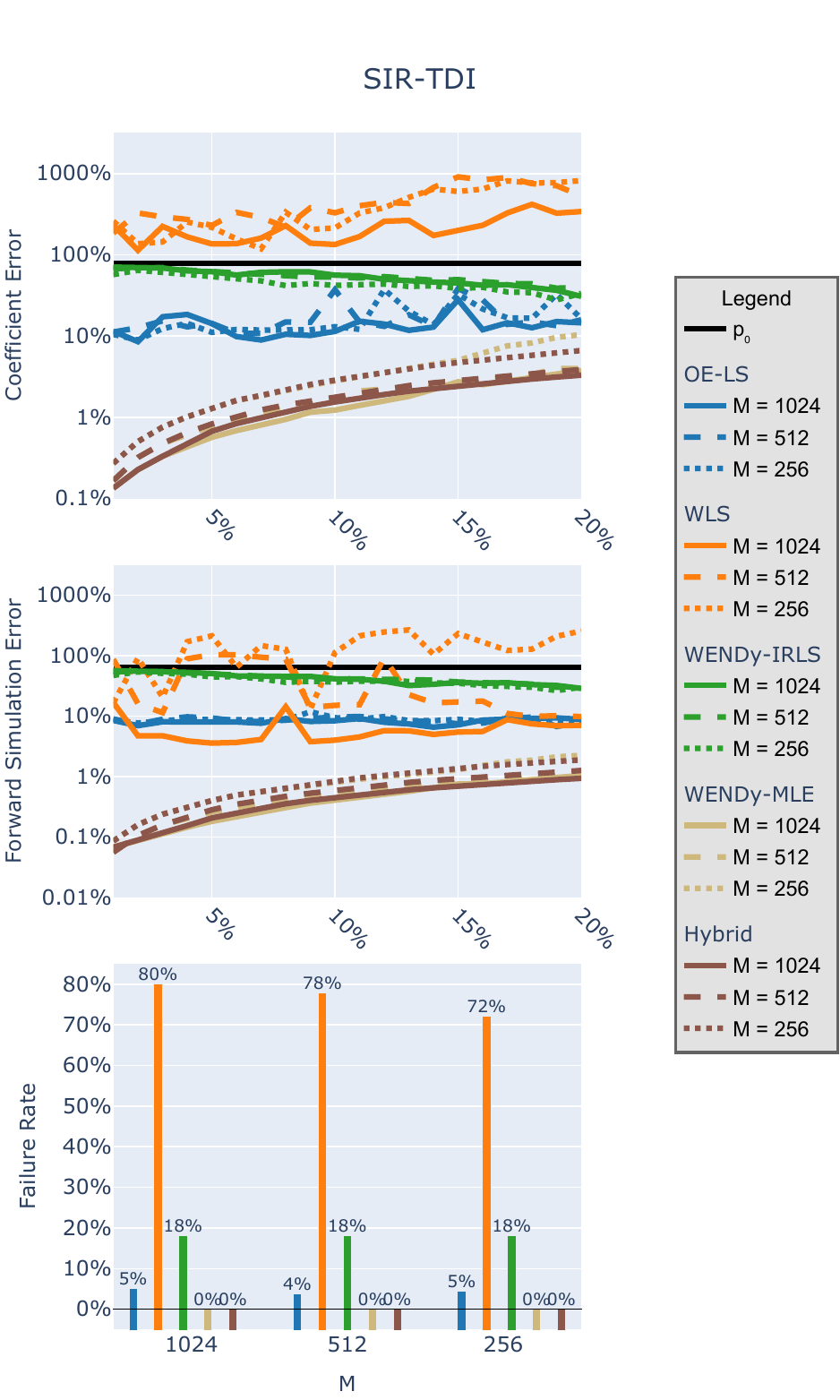}
	\caption{All algorithms are run at noise ratios from 1\% to 25\% and $M = \{256, 512, 1024\}$. We see accuracy metrics for for OE-LS in blue, WLS in orange, WENDy-IRLS in green, WENDy-MLE in gold, the hybrid method in brown, and initial parameterization in black.} 
	\label{fig:sirFleshedOut}
\end{figure}

\subsection{Estimator Plots}
The plots in Section \ref{appendix:confInt} give more detailed information each estimated parameter provided by WENDy-MLE and the provided covariance as obtained in Equation \ref{eq:param-dist}. This confidence intervals shown allow us to see when the true parameter value is within the 95\% confidence interval for each parameter.

The plots in Section \ref{appendix:estimQual} show estimated bias, variance, MSE and coverage for all parameters for the Hindmarsh-Rose, Goodwin and SIR-TDI examples. Each metric is plotted for each subsample rate with a different line specification. The metrics are plotted on the vertical axis vs the noise ratio on the horizontal axis.

\subsubsection{Confidence Intervals} \label{appendix:confInt}
\begin{figure}[H]
	\centering
    \includegraphics[width=1\textwidth]{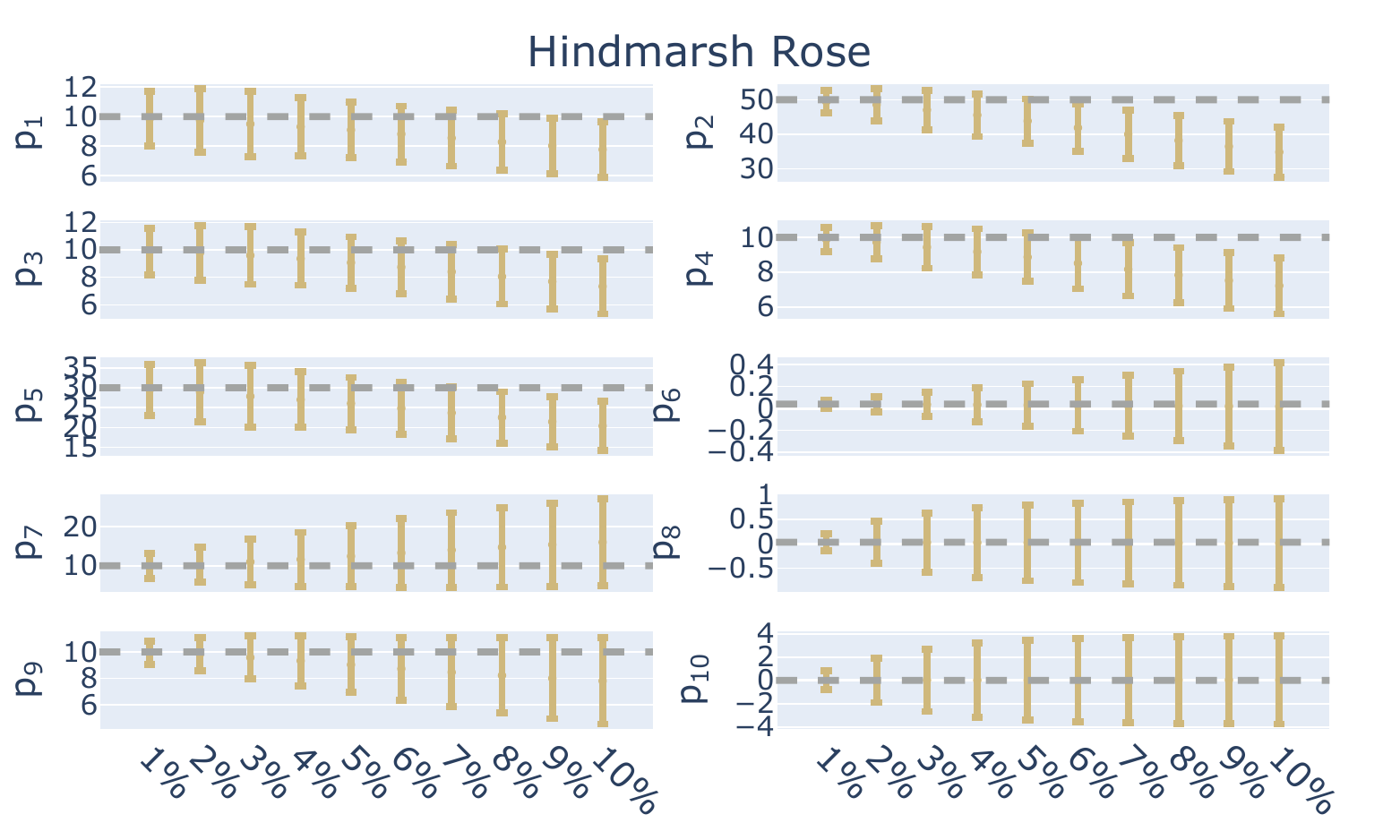}
	\caption{The WENDy-MLE's mean parameter estimate and mean 95\% confidence intervals for each parameter in gold and the true parameter is in gray.}
	\label{fig:hindmarshRose_confInt}
\end{figure}

\begin{figure}[H]
	\centering
	\includegraphics[width=1\textwidth]{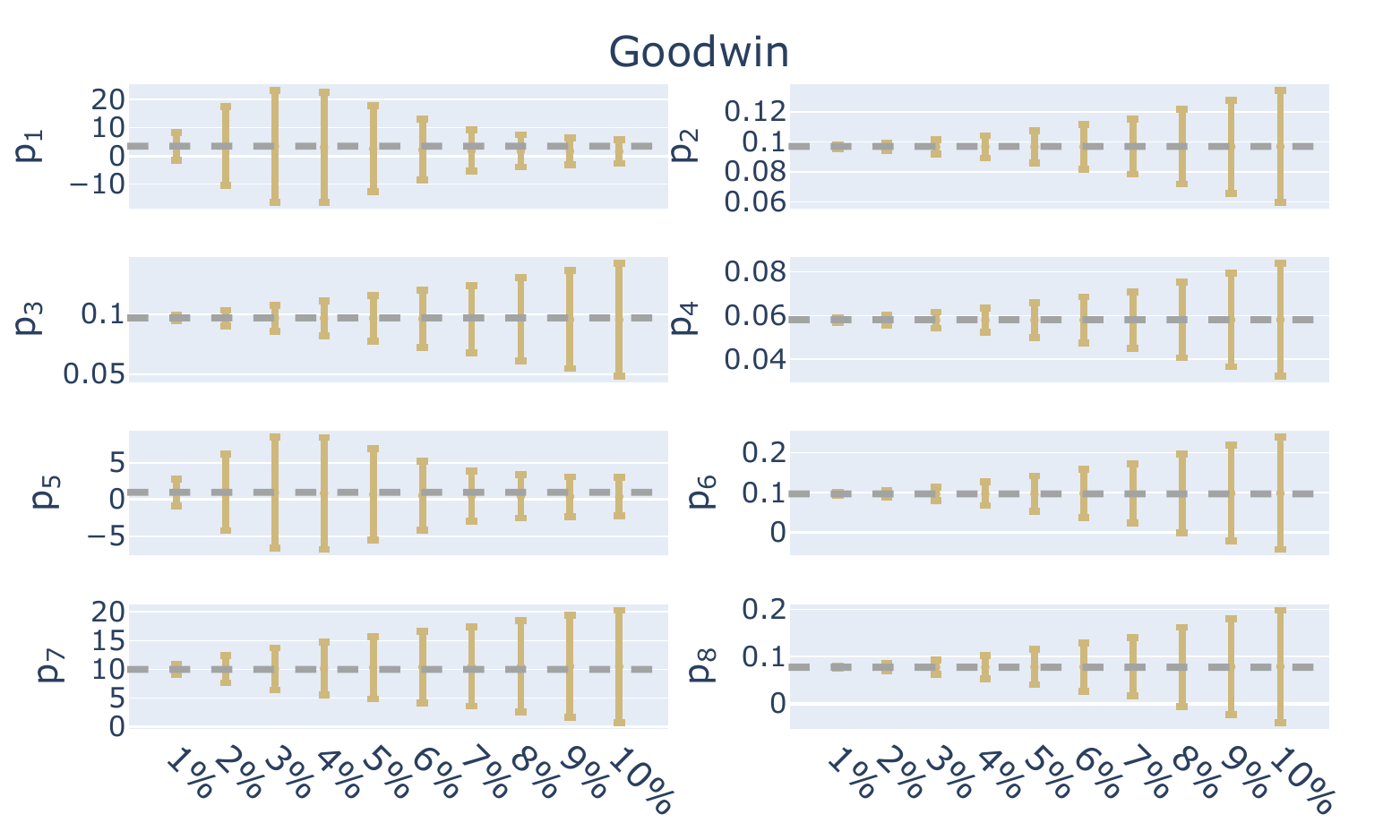}
	\caption{The WENDy-MLE's mean parameter estimate and mean 95\% confidence intervals for each parameter in gold and the true parameter is in gray.}
	\label{fig:goodwin_confInt}
\end{figure}

\begin{figure}[H]
	\centering
	\includegraphics[width=1\textwidth]{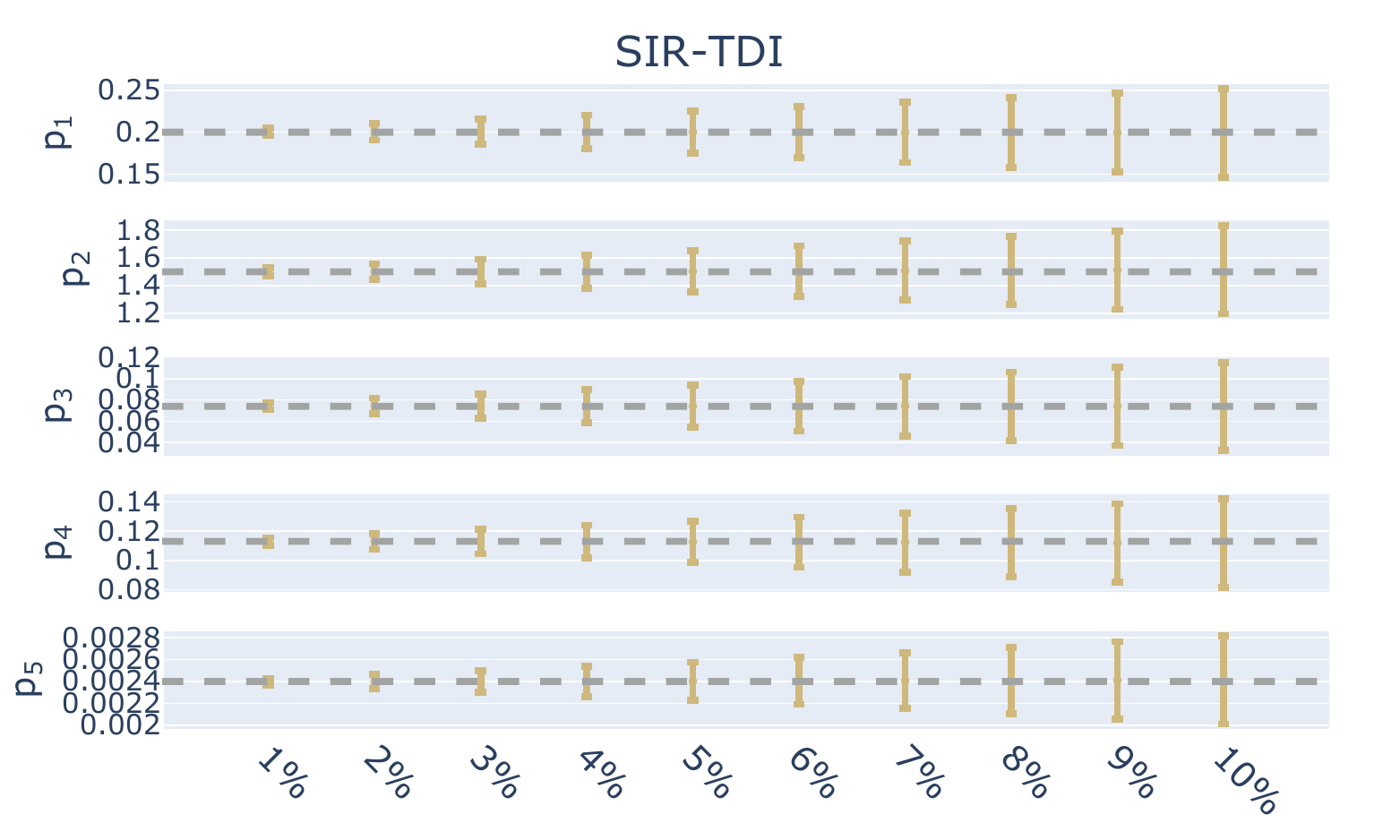}
	\caption{The WENDy-MLE's mean parameter estimate and mean 95\% confidence intervals for each parameter in gold and the true parameter is in gray.}
	\label{fig:sir_confInt}
\end{figure}

\newpage

\subsubsection{Estimator Quality Metrics}\label{appendix:estimQual}
\begin{figure}[H]
	\includegraphics[width=0.9\columnwidth]{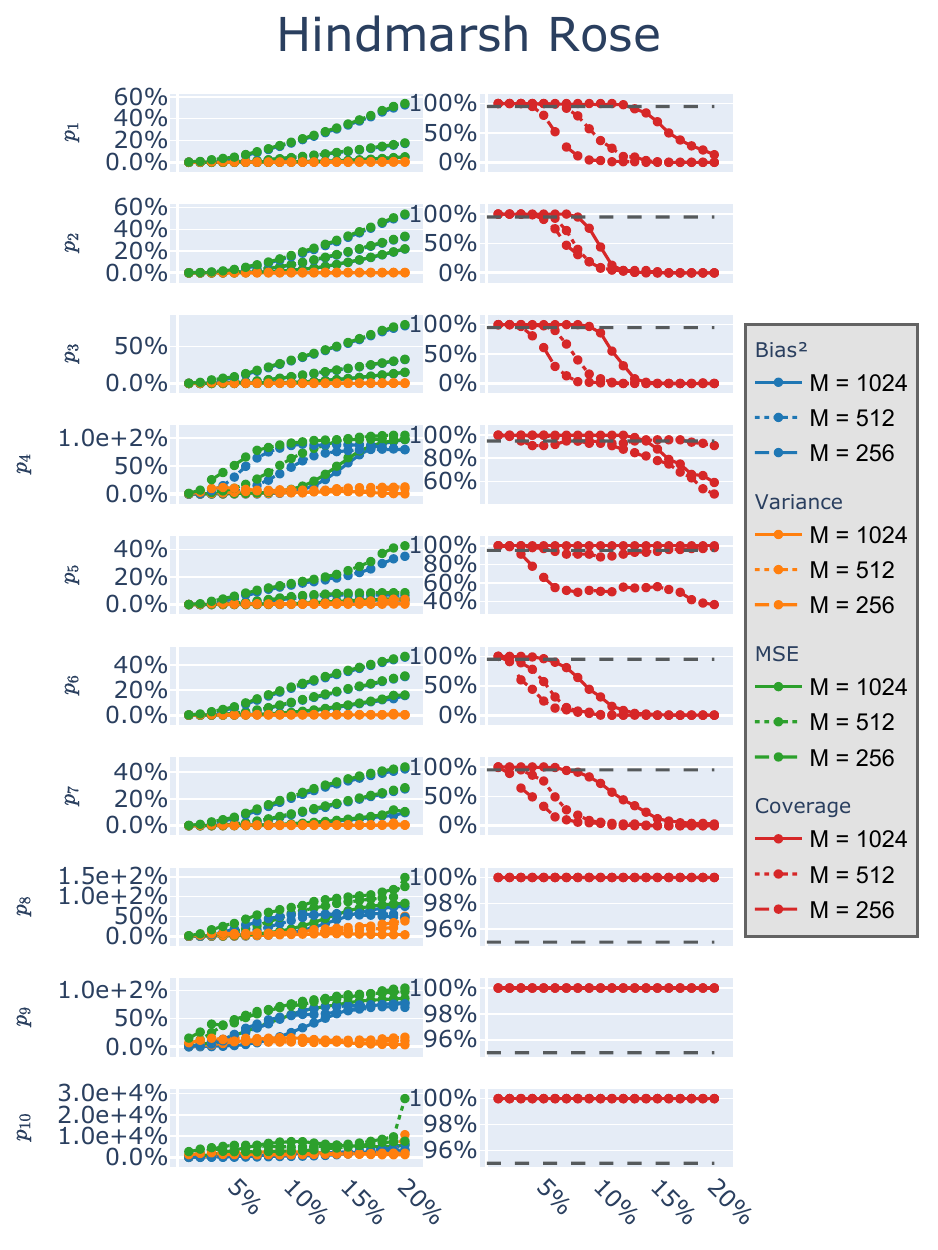}
	\caption{Left: the squared bias, variance and MSE for $p_1 - p_{10}$ top to bottom respectively, as a function of noise level. Right: the coverage levels for $p_1 - p_{10}$ top to bottom respectively, as a function of noise level.}
	\label{fig:hindmarshSup}
\end{figure}

\begin{figure}[H]
	\includegraphics[width=0.9\columnwidth]{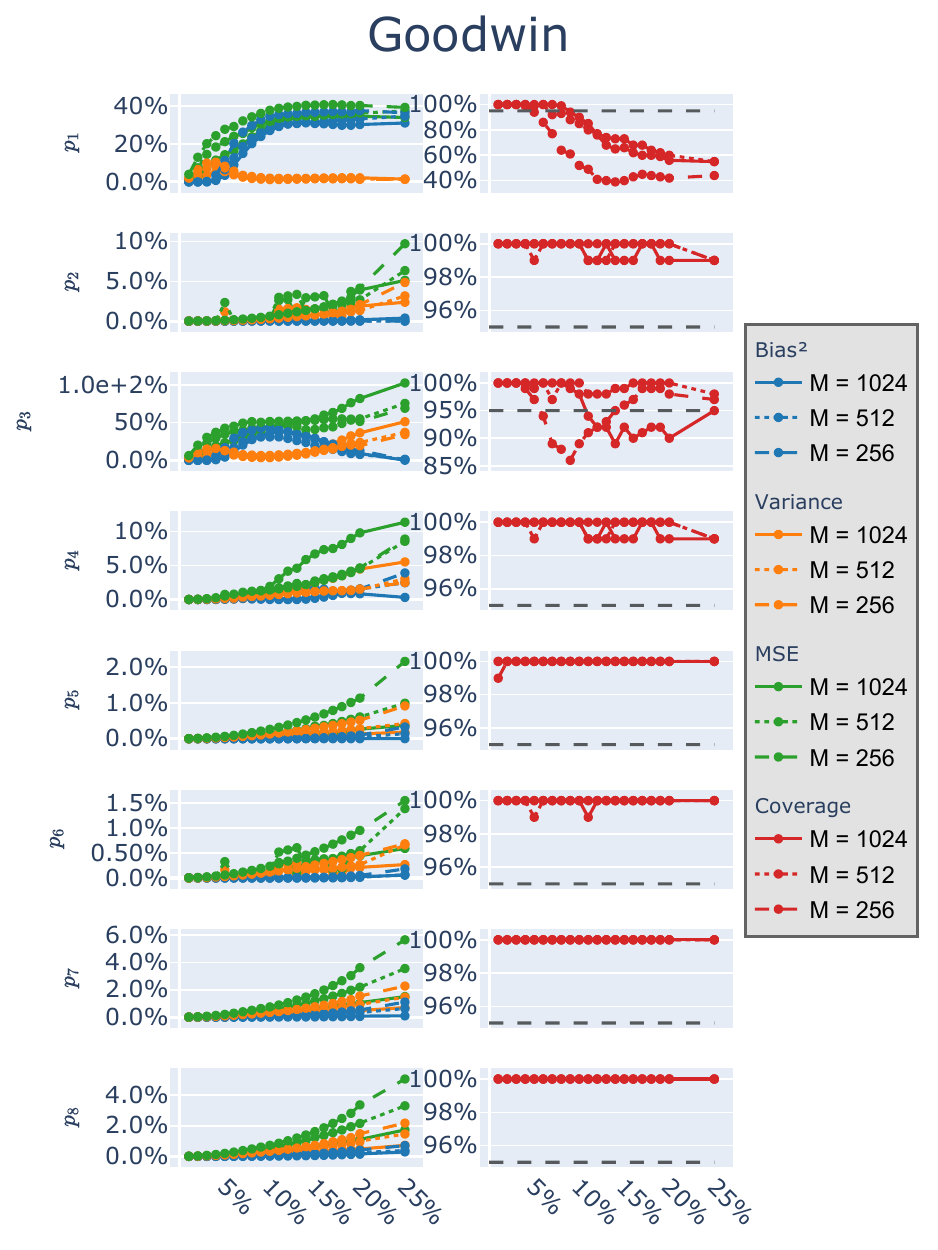}
	\caption{Left: the squared bias, variance and MSE for $p_1 - p8$ top to bottom respectively, as a function of noise level. Right: the coverage levels for $p_1 - p_8$ top to bottom respectively, as a function of noise level.}
	\label{fig:goodwinSup}
\end{figure}

\begin{figure}[H]
	\includegraphics[width=0.9\columnwidth]{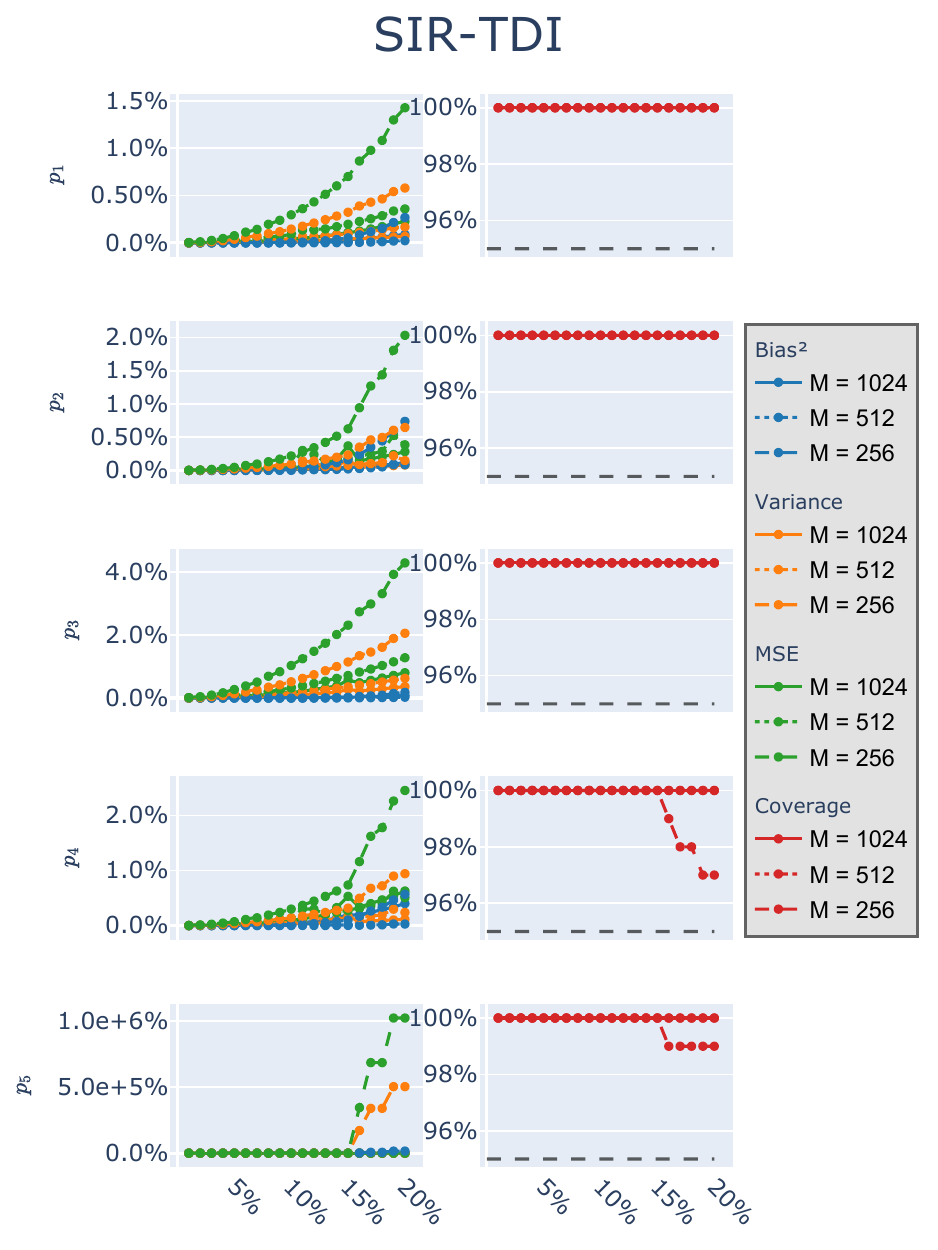}
	\caption{Left: the squared bias, variance and MSE for $p_1 - p_5$ top to bottom respectively, as a function of noise level. Right: the coverage levels for $p_1 - p_5$ top to bottom respectively, as a function of noise level.}
	\label{fig:sirSup}
\end{figure}
\end{document}